\documentclass[sigconf]{acmart}

\usepackage[utf8]{inputenc} % allow utf-8 input
\usepackage[T1]{fontenc}    % use 8-bit T1 fonts
\usepackage{hyperref}       % hyperlinks
\usepackage{url}            % simple URL typesetting
\usepackage{booktabs}       % professional-quality tables
\usepackage{amsfonts}       % blackboard math symbols
\usepackage{nicefrac}       % compact symbols for 1/2, etc.
\usepackage{microtype}      % microtypography

\usepackage{amsmath}
\usepackage{subcaption}
\usepackage{graphicx}
\usepackage{tikz}
\def\checkmark{\tikz\fill[scale=0.35](0,.35) -- (.25,0) -- (1,.7) -- (.25,.15) -- cycle;}
\usepackage{etoolbox}
\usepackage[ruled]{algorithm2e}
\usepackage{multirow}
\usepackage{hhline}
\usepackage{wrapfig}
\usepackage{comment}

\usepackage{xcolor}

\usepackage{rotating}

\usepackage{xspace}
\usepackage{units}

\newcommand{\mymath}[1]{\ensuremath{#1}\xspace}
\newcommand{\reals}{\mymath{\mathbb R}}

\definecolor{myred}{rgb}{0.8,0,0}
\definecolor{mygreen}{rgb}{0,0.6,0}
\definecolor{myblue}{rgb}{0,0,0.7}

\newcommand{\ddpg}{{\sc ddpg}\xspace}
\newcommand{\nslc}{{\sc nslc}\xspace}

\newcommand{\dvd}{{\sc d}v{\sc d}\xspace}

\newcommand{\mujoco}{{\sc MuJoCo}\xspace} 
\newcommand{\ant}{{\sc ant-v2}\xspace} 
\newcommand{\antmaze}{{\sc ant-maze}\xspace} 
\newcommand{\ptmaze}{{\sc point-maze}\xspace}
 
\newcommand{\anttrap}{{\sc ant-trap}\xspace}

\newcommand{\qdpg}{{\sc qd-pg}\xspace} 
\newcommand{\qpg}{{\sc q-pg}\xspace}
\newcommand{\dpg}{{\sc d-pg}\xspace}
\newcommand{\qdpgsum}{{\sc qd-pg sum}\xspace}

\newcommand{\cpu}{{\sc cpu}\xspace} 
 
\newcommand{\gpu}{{\sc gpu}\xspace} 
 
\newcommand{\geppg}{{\sc gep-pg}\xspace} 
\newcommand{\tddd}{{\sc td3}\xspace} 
\newcommand{\sac}{{\sc sac}\xspace} 
\newcommand{\cemrl}{{\sc cem-rl}\xspace} 
\newcommand{\cem}{{\sc cem}\xspace} 
\newcommand{\arac}{{\sc arac}\xspace} 
\newcommand{\pssstddd}{{\sc p3s-td3}\xspace}
\newcommand{\agac}{{\sc agac}\xspace}
\newcommand{\diayn}{{\sc diayn}\xspace}

\newcommand{\me}{{\sc map-elites}\xspace}
\newcommand{\cvtme}{{\sc cvt map-elites}\xspace}
\newcommand{\pgame}{{\sc pga-me}\xspace} 
\newcommand{\erl}{{\sc erl}\xspace} 
\newcommand{\cerl}{{\sc cerl}\xspace} 
\newcommand{\mees}{{\sc me-es}\xspace}

\newcommand{\qdes}{{\sc qd-es}\xspace} 
 
\newcommand{\nses}{{\sc ns-es}\xspace} 
\newcommand{\rnd}{{\sc rnd}\xspace}

\newcommand{\nsres}{{\sc nsr-es}\xspace} 
\newcommand{\nsraes}{{\sc nsra-es}\xspace} 
\newcommand{\goex}{{\sc Go-Explore}\xspace}

\newtheorem{proposition}{Proposition}

\AtBeginDocument{%
  \providecommand\BibTeX{{%
    \normalfont B\kern-0.5em{\scshape i\kern-0.25em b}\kern-0.8em\TeX}}}

\setcopyright{acmlicensed}
\acmConference[GECCO '22]{Genetic and Evolutionary Computation Conference}{July 9--13, 2022}{Boston, MA, USA}
\acmBooktitle{Genetic and Evolutionary Computation Conference (GECCO '22), July 9--13, 2022, Boston, MA, USA}
\acmPrice{15.00}
\copyrightyear{2022}
\acmYear{2022}
\acmDOI{10.1145/3512290.3528845}
\acmISBN{978-1-4503-9237-2/22/07}

\begin{document}

\title{Diversity Policy Gradient for Sample Efficient Quality-Diversity Optimization}


\author{Thomas Pierrot}
\authornote{Both authors contributed equally to this research.}
\affiliation{%
  \institution{InstaDeep}
  \city{Paris}
  \country{France}
}
\email{t.pierrot@instadeep.com}

\author{Valentin Mac\'e}
\authornotemark[1]
\affiliation{%
  \institution{InstaDeep}
  \city{Paris}
  \country{France}
}
\email{v.mace@instadeep.com}

\author{Felix Chalumeau}
\affiliation{%
  \institution{InstaDeep}
  \city{Paris}
  \country{France}
}
\email{f.chalumeau@instadeep.com}

\author{Arthur Flajolet}
\affiliation{%
  \institution{InstaDeep}
  \city{Paris}
  \country{France}
}
\email{a.flajolet@instadeep.com}

\author{Geoffrey Cideron}
\affiliation{%
  \institution{InstaDeep}
  \city{Paris}
  \country{France}
}
\email{geoffrey.cideron@gmail.com}

\author{Karim Beguir}
\affiliation{%
 \institution{InstaDeep}
 \city{London}
 \country{United Kingdom}
}
\email{kb@instadeep.com}

\author{Antoine Cully}
\affiliation{%
  \institution{Imperial College London}
  \city{London}
  \country{United Kingdom}
}
\email{a.cully@imperial.ac.uk}

\author{Olivier Sigaud}
\affiliation{%
  \institution{Sorbonne Universit\'e}
  \city{Paris}
  \country{France}
}
\email{olivier.sigaud@upmc.fr}

\author{Nicolas Perrin-Gilbert}
\affiliation{%
  \institution{CNRS, Sorbonne Universit\'e}
  \city{Paris}
  \country{France}
}
\email{perrin@isir.upmc.fr}

\renewcommand{\shortauthors}{Pierrot and Mac\'e, et al.}

\begin{abstract}
  A fascinating aspect of nature lies in its ability to produce a large and diverse collection of organisms that are all high-performing in their niche. By contrast, most AI algorithms focus on finding a single efficient solution to a given problem. Aiming for diversity in addition to performance is a convenient way to deal with the exploration-exploitation trade-off that plays a central role in learning. It also allows for increased robustness when the returned collection contains several working solutions to the considered problem, making it well-suited for real applications such as robotics. Quality-Diversity (QD) methods are evolutionary algorithms designed for this purpose. This paper proposes a novel algorithm, \qdpg, which combines the strength of Policy Gradient algorithms and Quality Diversity approaches to produce a collection of diverse and high-performing neural policies in continuous control environments. The main contribution of this work is the introduction of a Diversity Policy Gradient (DPG) that exploits information at the time-step level to drive policies towards more diversity in a sample-efficient manner. Specifically, \qdpg selects neural controllers from a \me grid and uses two gradient-based mutation operators to improve both quality and diversity. Our results demonstrate that \qdpg is significantly more sample-efficient than its evolutionary competitors.
\end{abstract}




%\ccsdesc[500]{Computer systems organization~Embedded systems}
%\ccsdesc{Computer systems %organization~Robotics}


\begin{teaserfigure}
    \begin{subfigure}{.6\linewidth}
        \includegraphics[width=\textwidth]{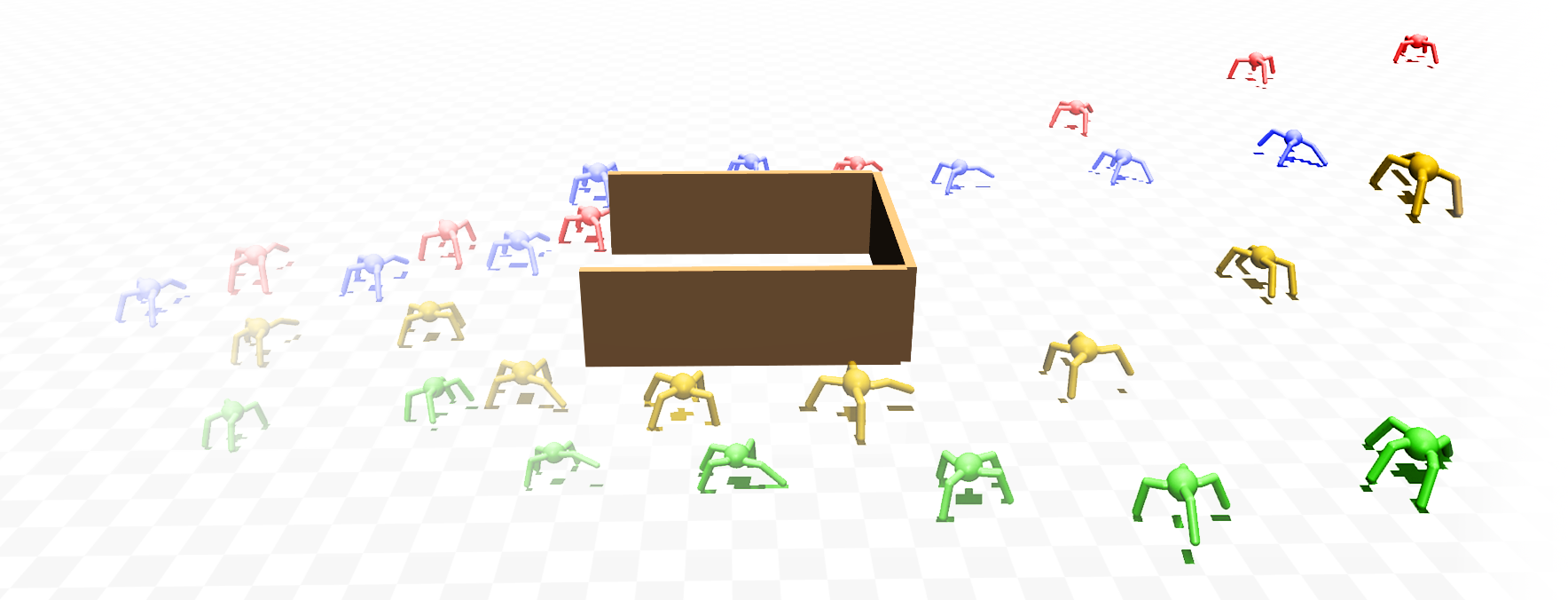}
        \Description{AntTrap final solutions}
    \end{subfigure}
    \hfill
    \begin{subfigure}{.4\linewidth}
        \includegraphics[width=\textwidth]{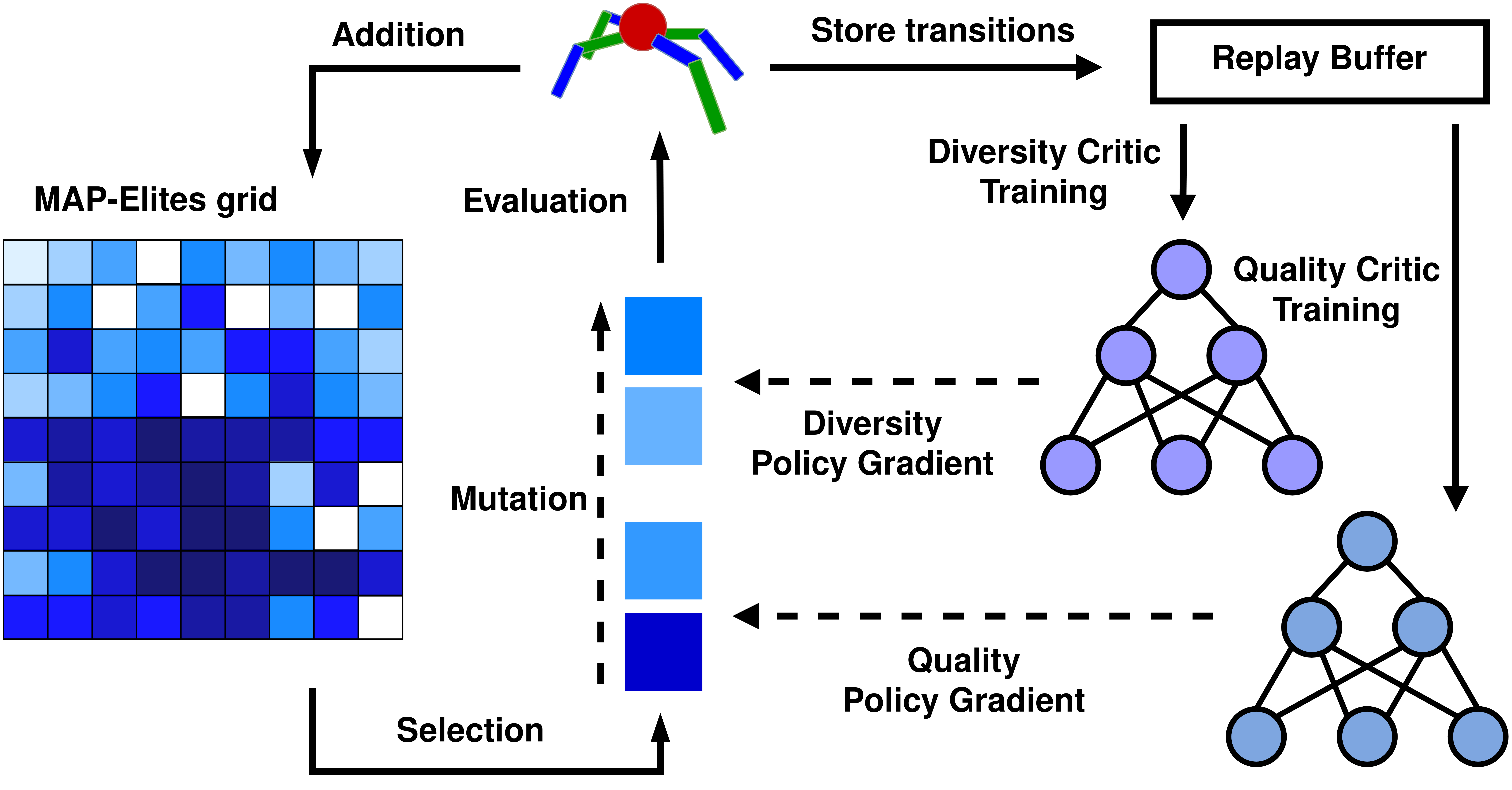}
        \Description{QDPG mechanism}
    \end{subfigure}
    \caption{An agent robot is rewarded for running forward as fast as possible. Following the reward signal without further exploration leads the agent into a trap corresponding to a poor local minimum. Our method \qdpg produces a collection of solutions that are diverse and high-performing, allowing a deeper exploration, necessary to solve deceptive control problems. \qdpg builds on the \me framework and leverages reinforcement learning  to derive policy gradient based mutations.
    }
    \label{fig:teaser}
\end{teaserfigure}

\maketitle

%\begin{figure}
%    \centering
%    \includegraphics[width=0.46\textwidth]{images/qdpg_schema.png}
%    \caption{Global mechanism of \qdpg algorithm. \qdpg builds upon the \me framework with selection, mutation, evaluation and addition steps. Instead of using a genetic operator during the mutation step, \qdpg leverages the data-efficiency of reinforcement learning by using a quality critic and a diversity critic to derive policy gradient based mutations in order to seek for performance and diversity.}
%    \label{fig:qdpg_bigpicture}
%\end{figure}

%\begin{figure}
%    \centering
%    \includegraphics[width=0.46\textwidth]{images/aliou_anttrap3_cropped.png}
%    \caption{The agent robot is rewarded for running forward as fast as possible. Following the reward signal without further exploration leads the agent into the trap, which corresponds to a poor local minimum. \qdpg produces a collection of solutions that are diverse and high-performing, allowing to find several working alternatives to solve a deceptive control problem.}
%    \label{fig:qdpg_bigpicture}
%\end{figure}

\section{Introduction}
Natural evolution has the fascinating ability to produce diverse organisms that are all well adapted to their respective niche. Inspired by this ability to produce a tremendous diversity of living systems, Quality-Diversity (QD) is a new family of optimization algorithms that aims at searching for a collection of both diverse and high-performing solutions \citep{pugh2016quality, cully2017quality}. While classic optimization methods focus on finding a single efficient solution, QD optimization aims to cover the range of possible solution types and to return the best solution for each type. This process is sometimes referred to as ``illumination" in opposition to optimization, as it reveals (or illuminates) a search space of interest often called the {\em behavior descriptor space} \citep{mouret2015illuminating}.

The principal advantage of QD approaches resides in their intrinsic capacity to deliver a large and diverse set of working alternatives when a single solution fails \citep{cully2015robots}. By producing a collection of solutions instead of a unique one, QD algorithms allow to obtain different ways to solve a single problem, leading to greater robustness, which can help to reduce the reality gap when applied to robotics \citep{koos2012transferability}.
Diversity seeking is the core component that allows QD algorithms to generate large collections of diverse solutions. By encouraging the emergence of novel behaviors in the population without focusing on performance alone, diversity seeking algorithms explore regions of the behavior descriptor space that are unreachable in practice for conventional algorithms \citep{doncieux2019novelty}.
Another benefit of QD is its ability to solve hard exploration problems where the reward signal is sparse or deceptive, and on which standard optimization techniques are ineffective \citep{colas2020scaling}. This ability can be interpreted as a direct consequence of the structured search for diversity in the behavior descriptor space.

Quality-Diversity algorithms build on black-box optimization methods such as evolutionary algorithms to evolve a population of solutions \citep{cully2017quality}. Historically, they rely on random mutations to explore small search spaces but struggle when facing higher-dimensional problems. As a result, they often scale poorly to problems where neural networks with many parameters provide state-of-the-art results \citep{colas2020scaling}. 

Building large and efficient controllers that work with continuous actions has been a long-standing goal in Artificial Intelligence and in particular in robotics. 
Deep reinforcement learning (RL), and especially Policy Gradient (PG) methods have proven efficient at training such large controllers \citep{schulman2017proximal, lillicrap2015continuous, fujimoto2018addressing, haarnoja2018soft}.
One of the keys to this success lies in the fact that PG methods exploit the structure of the objective function when the problem can be formalized as a Markov Decision Process (MDP), leading to substantial gains in sample efficiency. Moreover, they also exploit the analytical structure of the controller when known, which allows the sample complexity of these methods to be independent of the parameter space dimensionality \citep{vemula2019contrasting}. In real-world applications, these gains turn out to be critical when interacting with the environment is expensive.

Although exploration is very important to reach optimal policies, PG methods usually rely on simple exploration mechanisms, like adding Gaussian noise \citep{fujimoto2018addressing} or maximizing entropy \citep{haarnoja2018soft} to explore the action space, which happens to be insufficient in hard exploration tasks where the reward signal is sparse or deceptive \citep{colas2018gep, nasiriany2019planning}. Successful attempts have been made to combine evolutionary methods and reinforcement learning \citep{khadka2019collaborative, khadka2018evolutionaryNIPS, pourchot2018cem} to improve exploration. However, all these techniques only focus on building high-performing solutions and do not explicitly encourage diversity within the population. In this regard, they fail when confronted with hard exploration problems. More recently, Policy Gradient Assisted MAP-Elites (\pgame) \citep{Nilsson2021} was proposed to bring reinforcement learning in \me \citep{mouret2015illuminating} to train neural networks to solve locomotion tasks with diverse behaviors. Although using policy gradient for performance, the diversity seeking mechanism is still based on a divergent genetic search, making it struggle on hard exploration tasks.\\

\textbf{Contributions.} In this work, we introduce the idea of a {\em diversity policy gradient} (\dpg) that drives solutions towards more diversity. We show that the \dpg can be used in combination with the standard policy gradient, dubbed {\em quality policy gradient} (\qpg), to produce high-performing and diverse solutions. Our algorithm, called \qdpg, builds on \me and \pgame, replacing random diversity search by \dpg, and demonstrates remarkable sample efficiency brought by off-policy PG methods, and produces collections of good solutions in a single run (see \figurename~\ref{fig:teaser}).
We compare \qdpg to state-of-the-art policy gradient methods algorithms (including \sac, \tddd, \rnd, \agac, \diayn and \pgame), and to several evolutionary methods known as Evolution Strategies (ESs) augmented with a diversity objective (namely the \nses family \citep{conti2018improving} and the \mees algorithm \citep{colas2020scaling}) on a set of challenging control tasks with deceptive reward signal.

We show that \qdpg generates collections of robust and high-performing solutions in hard exploration problems while standard policy gradient algorithms struggle to produce a single one. \qdpg is several order of magnitude more sample efficient than its traditional evolutionary competitors and outperforms \pgame on all benchmarks.

\section{Background}

\subsection{Problem statement}
\label{sec:problem_statement}

We consider an MDP $\left(\mathcal{S}, \mathcal{A}, \mathcal{R}, \mathcal{T}, \gamma \right)$ where $\mathcal{S}$ is the state space, $\mathcal{A}$ the action space, $\mathcal{R} : \mathcal{S} \times \mathcal{A} \rightarrow \mathbb{R}$ the reward function, $\mathcal{T}: \mathcal{S} \times \mathcal{A} \rightarrow \mathcal{S}$  the dynamics transition function and $\gamma$ a discount factor. We assume that both $\mathcal{S}$ and $\mathcal{A}$ are continuous and consider a controller, or policy, $\pi_{\theta}: \mathcal{S} \rightarrow \mathcal{A}$, a neural network parameterized by $\mathbf{\theta} \in \Theta$, which is called a {\em solution} to the problem. 
The \emph{fitness} $F: \Theta \rightarrow \mathbb{R}$ of a solution measures its performance, defined as the expectation over the sum of rewards obtained by controller $\pi_{\theta}$. A solution with high fitness is said to be {\em high-performing}.

We introduce a behavior descriptor (BD) space $\mathcal{B}$, a behavior descriptor extraction function $\mathbf{\xi}: \Theta \rightarrow \mathcal{B}$, and define a distance metric $||.||_{\mathcal{B}}$ over $\mathcal{B}$. The {\em diversity} of a set of $K$ solutions $\{\theta_k\}_{k=1,\dots,K}$ is defined as $d: \Theta^K \rightarrow \mathbb{R}^+$:

\begin{equation}
\label{eq:diversity_computation}
    d\left(\{\mathbf{\theta}_k\}_{k=1,\dots,K}\right) = \sum\limits_{i=1}^K \min\limits_{k \ne i} ||\mathbf{\xi}(\mathbf{\theta}_i), \mathbf{\xi}(\mathbf{\theta}_k)||_{\mathcal{B}},
\end{equation}
meaning that a set of solutions is diverse if the solutions are distant with respect to each other in the sense of $||.||_{\mathcal{B}}$.

Following the objective of the family of Quality Diversity algorithms, we are trying to evolve a population of {\em diverse} and {\em high-performing} solutions.

\subsection{The MAP-Elites algorithm}

\me \citep{mouret2015illuminating} is a simple yet state-of-the-art QD algorithm that has been successfully applied to a wide range of challenging problems such as robot damage recovery \citep{cully2015robots}, molecular robotic control \citep{cazenille2019exploring} and game design \citep{alvarez2019empowering}.
In \me, the behavior descriptor space $\mathcal{B}$ is discretized into a grid of cells, also called niches, with the aim of filling each cell with a high-performing solution. A variant, called \cvtme uses Centroidal Voronoi Tesselations \citep{DBLP:journals/corr/VassiliadesCM16} to initially divide the grid into the desired number of cells. The algorithm starts with an empty grid and an initial random set of $K$ solutions that are evaluated and added to the grid by following simple insertion rules. If the cell corresponding to the behavior descriptors of a solution is empty, then the solution is added to this cell. If there is already a solution in the cell, the new solution replaces it only if it has greater fitness. At each iteration, $P$ existing solutions are sampled uniformly from the grid and randomly mutated to create $P$ new solutions. These new solutions are then evaluated and added to the grid following the same insertion rules. This cycle is repeated until convergence or for a given budget of iterations. \me is a compelling and efficient method. However, it suffers from a low sample efficiency as it relies on random mutations.

\subsection{Reinforcement Learning and TD3}
\label{sec:reinforcementlearning}

Deep Reinforcement learning (DRL) is a paradigm to learn high-performing policies implemented by neural networks in MDPs. In this work, we focus on a class of policy search methods called policy gradient methods. In opposition to standard evolutionary methods that rely on random updates, policy gradient methods exploit the structure of the MDP under the form of the Bellman equations to compute efficient performance-driven updates to improve the policy. 

Among many algorithms in the reinforcement learning community, \tddd \citep{fujimoto2018addressing} shows state-of-the-art performance to train controllers in environments with continuous action space and large state space. \tddd relies on the deterministic policy gradient update \cite{silver2014deterministic} to train deterministic policies, as defined in Section \ref{sec:problem_statement}. In most policy gradients methods, a critic $Q^\pi: \mathcal{S} \times \mathcal{A} \rightarrow \mathbb{R}$ implemented by a neural network is introduced. The critic evaluates the expected fitness of policy $\pi$ when performing a rollout starting from a state-action pair $(s, a)$. The policy is updated to take the actions that will maximise the critic's value estimation in each state. Both the actor and the critic have an associated target network and use delayed policy updates to stabilize the training.

The \tddd algorithm adds on top of the deterministic policy gradient update additional mechanisms that were demonstrated to significantly improve performance and stability in practice \citep{fujimoto2018addressing}. Namely, two critics are used together and the minimum of their values is used to mitigate overestimation issues. A Gaussian noise is added to the actions taken by the actor when computing the deterministic policy gradient to enforce smoothness between similar action values. More details can be found in the supplementary material in Section \ref{sec:td3}.

\subsection{Policy Gradient Assisted MAP-Elites}
\label{sec:pgame}

Policy Gradient Assisted MAP-Elites (\pgame, \citep{Nilsson2021}) is a recent QD algorithm designed to evolve populations of neural networks. \pgame builds on top of the \me algorithm and uses policy gradient updates to create a mutation operator that drives the population towards region of high fitness. Used along genetic mutations, it has shown the ability to evolve high-performing neural networks controllers where \me with genetic mutations alone failed. For this purpose, \pgame introduces a replay buffer that stores all the transitions experienced by controllers from the population during evaluation steps and an actor/critic couple that uses those stored transitions to train, following the \tddd method presented in \ref{sec:reinforcementlearning}. The critic is used at each iteration to derive the quality policy gradient estimate for half of the offsprings selected from the \me grid. The critic is trained with a dedicated actor, called the greedy controller. This greedy controller is also updated at each iteration by the critic and is added to the \me grid when relevant. This greedy controller is never discarded, even if its fitness is lower than individuals of similar behavior. 

\section{Key Principle: Diversity Policy Gradient}
\label{sec:key}

\subsection{General principle}
In this work, we introduce the Diversity Policy Gradient, which aims at using time-step level information in order to mutate controllers towards diversity. We are trying to characterize the behavior of policies as a function of the states they visit and hence construct a diversity reward at the time-step-level that is correlated with the diversity at the episode-level. This reward is computed based on distance of a state compared to the nearest states visited by other controllers in the \me grid. Maximizing the accumulated diversity rewards will induce an increase of the diversity in the population we are optimizing. The following section provides a mathematical motivation for this approach.

\subsection{Mathematical formulation}
In this section, we motivate and introduce formally the \dpg computations.

Let us assume that we have a \me grid containing $K$ solutions $(\theta_1, \dots, \theta_K)$ and that we sampled $\theta_1$ from the grid to evolve it. We want to update $\theta_1$ in such a way that the population's diversity as defined in Equation~\eqref{eq:diversity_computation} will increase. For this purpose, we aim to compute the gradient of the diversity with respect to $\theta_1$ and update $\theta_1$ in its direction using standard gradient ascent techniques. 

\begin{proposition}
\label{prop:div_gradient_1}
The gradient of diversity with respect to $\theta_1$ can be written as
\begin{align}
    \begin{cases}
    \nabla_{\theta_1} d(\{\theta_k\}_{k=1,\dots,K}) = \nabla_{\theta_1} n(\theta_1, \left(\theta_j\right)_{ 2 \leq j \leq J}),\\
    \text{where} \ \ n(\theta_1, \left(\theta_j\right)_{ 2 \leq j \leq J}) = \sum\nolimits_{j=2}^J ||\xi(\theta_1), \xi(\theta_j)||_{\mathcal{B}}
    \end{cases}
\end{align} 
and $\theta_2$ is $ \theta_1$ closest neighbour and $\left(\theta_j\right)$ with $j = 3, \dots, J$ are the elements in the population for which $\theta_1$ is the nearest neighbour. Proof in Appendix~\ref{sec:proof}.
\end{proposition}

We call $n$ the novelty of $\theta_1$ with respect to its nearest elements. This proposition means that we can increase the diversity of the population by increasing the novelty of $\theta_1$ with respect to the solutions for which it is the nearest neighbor.

Under this form, the diversity gradient cannot benefit from the variance reduction methods in the RL literature to efficiently compute policy gradients \cite{sutton1999policy}. To this end, we need to express it as a gradient over the expectation of a sum of scalar quantities obtained by policy $\pi_{\theta_1}$ at each step when interacting with the environment.

For this purpose, we introduce a novel space $\mathcal{D}$, dubbed {\em state descriptor space} and a {\em state descriptor extraction function} $\psi: \mathcal{S} \rightarrow \mathcal{D}$. We assume $\mathcal{D}$ and $\mathcal{B}$ have the same dimension. The notion of state descriptor will be used in the following to link diversity at the time step level to the diversity at the trajectory level.

In this context, if the following compatibility equation is satisfied:

\begin{align}
\label{eq:diversity_gradient_constraint}
    \begin{cases}
     n(\theta_1, \left(\theta_j\right)_{ 2 \leq j \leq J}) = \mathbb{E}_{\pi_{\theta_1}} \sum\limits_t n(s_t, \left(\theta_j\right)_{ 2 \leq j \leq J}) \\
     \text{where} \ n(s, \left(\theta_j\right)_{j=1,\dots,J}) = \sum ^J\limits_{j=1} \mathbb{E}_{\pi_{\theta_{j}}} \sum\limits_t ||\psi(s), \psi(s_t)||_{\mathcal{D}}
     \end{cases}
\end{align}
then the diversity policy gradient can be computed as:

\begin{align}
    \label{eq:dpg}
    \nabla_{\theta_1} d(\{\theta_k\}_{k=1,\dots,K}) = \nabla_{\theta_1} \mathbb{E}_{\pi_{\theta_1}} \sum\limits_t n(s_t, \left(\theta_j\right)_{ 2 \leq j \leq J})
\end{align}

See the Appendix~\ref{sec:proof} for more details about the motivation behind this assumption. The obtained expression corresponds to the classical policy gradient setting where $\gamma = 1$ and where the corresponding reward signal, here dubbed diversity reward, is computed as $r^D_t = n(s_t, \left(\theta_j\right)_{\ 2 \leq j \leq J})$. Therefore, this gradient can be computed using any PG estimation technique replacing the environment reward by the diversity reward $r^D_t$.

Equation \eqref{eq:diversity_gradient_constraint} enforces a relation between $\mathcal{B}$ and $\mathcal{D}$ and between extraction functions $\psi$ and $\xi$. In practice, it may be hard to define the behavior descriptor and state descriptor of a solution that satisfy this relation while being meaningful to the problem at hand and tractable. Nevertheless, a strict equality is not necessary: a positive correlation between the two hand-sides of the equation is sufficient for diversity seeking.

\section{Related Work}
\label{sec:literature_review}

Simultaneously maximizing diversity and performance is the central goal of QD methods \citep{pugh2016quality,cully2017quality}. Among the various possible combinations offered by the QD framework \citep{cully2017quality}, Novelty Search with Local Competition (\nslc) \citep{lehman2011evolving} and \me \citep{mouret2015illuminating} are the two most popular algorithms. \nslc builds on the Novelty Search (NS) algorithm \citep{lehman2011abandoning} and maintains an unstructured archive of solutions selected for their local performance while \me uniformly samples individuals from a structured grid that discretizes the BD space. Although very efficient in small parameter spaces, those methods struggle to scale to bigger spaces which limits their application to neuro-evolution.\\

\textbf{Gradients in QD.} Hence, algorithms mixing quality-diversity and evolution strategies, dubbed \qdes, have been developed with the objective of improving the data-efficiency of the mutations used in QD thanks to Evolution Strategies. Algorithms such as \nsres and \nsraes have been applied to challenging continuous control environments \cite{conti2018improving}. But, as outlined by \citet{colas2020scaling}, they still suffer from poor sample efficiency and the diversity and environment reward functions could be mixed in a more efficient way. \mees \citep{colas2020scaling} went one step further in that direction, achieving a better mix of quality and diversity seeking through the use of a \me grid and two specialized ES populations. Using these methods was shown to be critically more successful than population-based GA algorithms \citep{salimans2017evolution}, but they rely on heavy computation and remain significantly less sample efficient than off-policy deep RL methods, as they do not leverage the analytical computation of the policy gradient at the time step level. Differentiable QD \citep{Fontaine2021} improves data-efficiency in QD with an efficient search in the descriptor space but does not tackle neuro-evolution and is limited to problems where the fitness and behavior descriptor functions are differentiable, which is not the case in a majority of robotic control environments.\\

\textbf{Exploration and diversity in RL.} Although very efficient to tackle large state/action space problems, most reinforcement learning algorithms struggle when facing hard exploration environments. Therefore, many efforts have been made to try to tackle it.
RL methods generally seek diversity either in the state space or in the action space. This is the case of algorithms maintaining a population of RL agents for exploration without an explicit diversity criterion \citep{jaderberg2017population} or algorithms explicitly looking for diversity but in the action space rather than in the state space like \arac \citep{doan2019attraction}, \agac \citep{AGAC2021}, \pssstddd \citep{jung2020population} and \dvd \citep{parker2020effective}.

An exception is \textit{Curiosity Search} \cite{stanton2016curiosity} which defines a notion of {\em intra-life novelty} that is similar to our state novelty defined in Section~\ref{sec:key}.
Our work is also related to algorithms using RL mechanisms to search for diversity only like \diayn \citep{eysenbach2018diversity} and others \citep{pong2019skew,lee2019efficient,islam2019marginalized}. These methods have proven useful in sparse reward situations, but they are inherently limited when the reward signal can orient exploration, as they ignore it. Other works sequentially combine diversity seeking and RL. The \geppg algorithm \cite{colas2018gep} combines a diversity seeking component, namely {\em Goal Exploration Processes} \citep{forestier2017intrinsically} and the \ddpg RL algorithm \citep{lillicrap2015continuous}. This combination of exploration-then-exploitation is also present in \goex \citep{ecoffet2019go}. These sequential approaches first look for diverse behaviors before optimizing performance.\\

\textbf{Mixing policy gradient and evolution.} The fruitful synergy between evolutive and RL methods has been explored in many recent methods (notably \erl \citep{khadka2018evolutionaryNIPS}, \cerl \citep{CERL}, \cemrl \citep{pourchot2018cem} and \pgame \citep{Nilsson2021}). \erl and \cemrl mix Evolution Strategies and RL to evolve a population of agents to maximize quality but ignores the diversity of the population. 

Policy Gradient Assisted MAP-Elites (\pgame) successfully combines QD and RL. This algorithm scales \me to neuroevolution by evolving half of its offsprings with a quality policy gradient update instead of using a genetic mutation alone. Nevertheless, the search for diversity is only explicitly done with the genetic mutation. Although based on a efficient crossover strategy \citep{Vassiliades2018}, it remains a divergent search with limited data-efficiency and ignores the analytical structure of the controller. Despite having been proven effective in classic locomotion environments, our experiments show that \pgame struggles on exploration environments with deceptive rewards. \qdpg builds on the ideas introduced in \pgame but replaces the genetic mutation by the diversity policy gradient introduced in section \ref{sec:key}. This data-efficient mechanism explicitly using time-step information to seek diversity helps scaling \me to neuroevolution for hard-exploration tasks.

To the best of our knowledge, \qdpg is the first algorithm optimizing both diversity and performance in the solution space and in the state space, using a sample-efficient policy gradient computation method for the latter.

\section{Methods}
\label{sec:methods}

\makeatletter
\makeatother

\SetKwComment{Comment}{/* }{ */}

\SetArgSty{textnormal}

\begin{algorithm}
    \small
    \SetAlgoLined
    \DontPrintSemicolon
    \SetKwInput{KwInput}{Given}
    \KwInput{
    \begin{itemize}
        \item Sample size N, Total number of steps max\_steps
        \item Behavior and state descriptor extraction functions $\xi$, $\psi$
        \item \me grid $\mathbb{M}$, state descriptors archive $\mathbb{A}$, Replay Buffer $\mathbb{B}$
        \item $N$ initial actors $\{\pi_{\theta_{i}}\}_{i=\{1,N\}}$ and
    2 critics $Q^D_w$, $Q^Q_v$\;
    \end{itemize}
    }
    \texttt{\\}

    \tcp{Main loop}
    $\textrm{total\_steps} \leftarrow 0$\;
    \While{$\textrm{total\_steps} < \textrm{max\_steps}$}{
    \texttt{\\}
    
        \If{$\textrm{total\_steps} > 0$}{
            \texttt{\\}
            \tcp{Sampling and mutation}
            Sample generation $\{\pi_{\theta_i}\}_{i = 1,N}$ in grid $\mathbb{M}$\;
            Sample batches of transitions in replay buffer $\mathbb{B}$\;
            Compute fresh novelty rewards for half the batches\;
            Update half the generation for diversity\;
            Update the other half for quality\;
            Update diversity and quality critics $Q^D_w$ and $Q^Q_v$\;
            
        }
        \texttt{\\}
        \tcp{Evaluation and insertion in grid}
        Evaluate the generation and store collected transitions in $\mathbb{B}$\;
        Store state descriptors in archive $\mathbb{A}$\;
        Update total\_steps with the total number of collected steps\;
        Add the updated generation in the grid $\mathbb{M}$\; 
    }

    \caption{QD-PG}
    \label{alg:compact_pseudocode}
\end{algorithm}

Our full algorithm is called \qdpg, its pseudo code is given in Algorithm~\ref{alg:compact_pseudocode} (full version and architecture in Appendix~\ref{sec:algo_supp}).
\qdpg is an iterative algorithm based on \me that replaces random mutations with policy gradient updates. As we consider a continuous action space and want to improve sample efficiency by using an off-policy policy gradient method, we rely on \tddd to compute policy gradients updates as in \pgame.

\qdpg maintains three permanent structures. On the QD side, a \cvtme grid stores the most novel and performing solutions. On the RL side, a replay buffer contains all transitions collected when evaluating solutions and an archive $\mathbb{A}$ stores all state descriptors obtained so far. \qdpg starts with an initial population of random solutions, evaluates them and inserts them into the \me grid. At each iteration, solutions are sampled from the grid, copied, and updated. The updated solutions are then evaluated through one rollout in the environment and inserted into the grid according to the usual insertion rules. Transitions collected during evaluation are stored in the replay buffer, and state descriptors are stored in the archive $\mathbb{A}$. Note that these state descriptors are first filtered to avoid insertion in the archive of multiple state descriptors that are too close to each other.

During the update step, half the population is updated with \qpg ascent and the other half with \dpg ascent. The choice of whether an agent is updated for quality or diversity is random, meaning that it can be updated for quality and later for diversity if selected again. To justify this design, we show in Section~\ref{sec:results} that updating consecutively for quality and diversity outperforms updating based on joint criteria. Both gradients are computed from batches of transitions sampled from the replay buffer. The \qpg is computed from the usual environment rewards (similar to \tddd or \pgame) whereas for \dpg, we get "fresh" novelty rewards as
\begin{equation}
    r_t^D = \sum^J\limits_{j=1} ||\psi(s_t), \psi(s_j)||_{\mathcal{D}},
\end{equation}
where $\left(s_j\right)_{j=1,\dots,J}$ are the $J$ nearest neighbors of state $s_t$ in the archive $\mathbb{A}$. Diversity rewards must be recomputed at each update because $\mathbb{A}$ changes during training. Following Equation~\eqref{eq:diversity_gradient_constraint}, diversity rewards should be computed as the sum of the distances between the descriptor of $s_t$ and the descriptors of all the states visited by a list of $J$ solutions.  In practice, we consider the $J$ nearest neighbors of $s_t$. This choice simplifies the algorithm, is faster to compute and works well in practice. 

\tddd relies on a parameterized critic to reduce the variance of its policy gradient estimate. In \qdpg, we maintain two parameterized critics $Q^D_{w}$ and $Q^Q_{v}$, respectively dubbed diversity and quality critics. Every time a policy gradient is computed, \qdpg also updates the corresponding critic. The critics are hence trained with all agents in the population instead of a specialised agent (like in \pgame). This helps avoiding local minima in exploration environments where the specialised actor could get stuck and hence mislead the values learned by the critic. However, having critics trained with multiple agents can destabilize the process, which is why we avoid using \qdpg with big grids. In our benchmarks, our grids usually contain 3 times less cells than \pgame.
As in \tddd, we use pairs of critics and target critics to fight the overestimation bias. We share the critic parameters among the population as in \cem \cite{pourchot2018cem}. Reasons for doing so come from the fact that diversity is not stationary, as it depends on the current population. If each agent had its own diversity critic, since an agent may not be selected for a large number of generations before being selected again, its critic would convey an outdated picture of the evolving diversity. We tried this solution, and it failed. 
A side benefit of critic sharing is that both critics become accurate faster as they combine experience from all agents. Additional details on \qdpg implementation are available in Appendix~\ref{sec:qdrl_details}.

\section{Experiments}
\label{sec:exps}

In this section, we intend to answer the following questions: 1. Can \qdpg produce collections of diverse and high-performing neural policies and what are the advantages to do so? 2. Is \qdpg more sample efficient than its evolutionary competitors? 3. How difficult are the considered benchmarks for classical policy gradients methods? 4. What is the usefulness of the different components of \qdpg ? 
 
\subsection{Environments}

We assess \qdpg capabilities in continuous control environments that exhibit high dimensional observation and action spaces as well as deceptive rewards. The size of the state/action space makes exploration difficult for Genetic Algorithms and Evolution Strategies. Deceptive rewards creates exploration difficulties which is particularly challenging for classical RL methods. We consider three OpenAI Gym environments based on the \mujoco physics engine that all exhibit strong deceptive rewards (illustrated in the Appendix in Figure~\ref{fig:grad_maps}), \ptmaze, \antmaze and \anttrap. Such environments have also been widely used in previous works \citep{parker2020effective, colas2020scaling, frans2017meta, shi2020efficient}.

\begin{figure}[h!] %[thbp!]
 \centering
 \begin{subfigure}{.32\linewidth}
 \centering
   \includegraphics[width=1\linewidth]{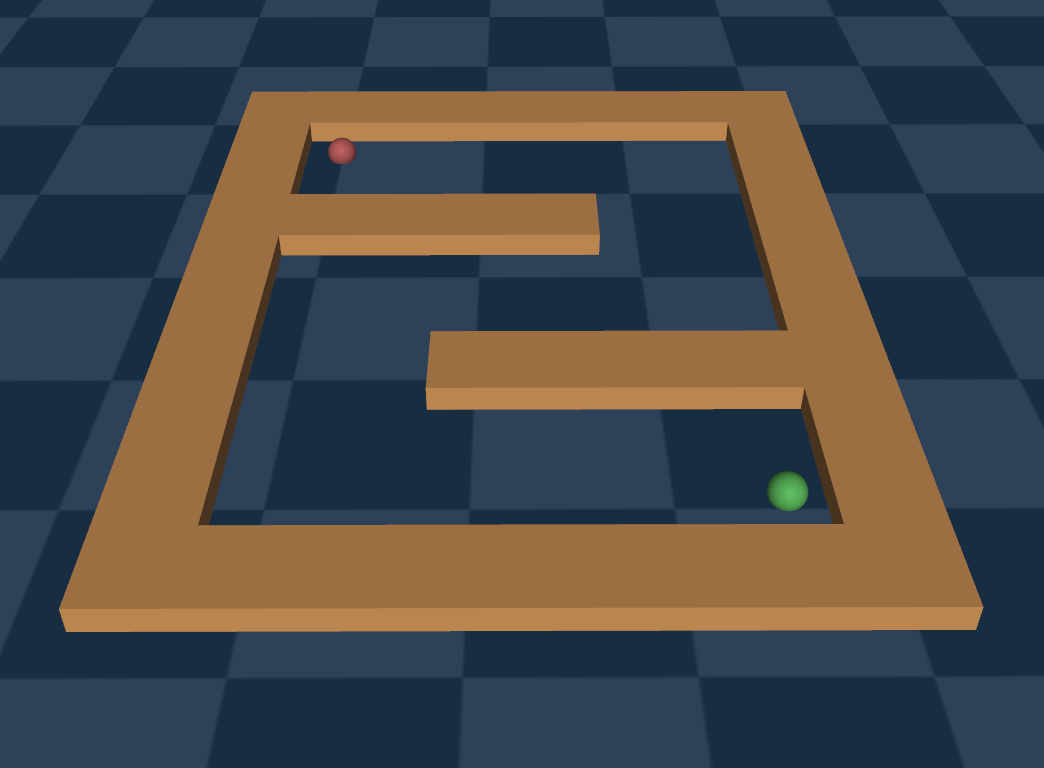}
   \caption{\ptmaze}
   \label{fig:point_maze_inertia}
 \end{subfigure}%
 \hfill
 \begin{subfigure}{.32\linewidth}
 \centering
   \includegraphics[width=1\linewidth]{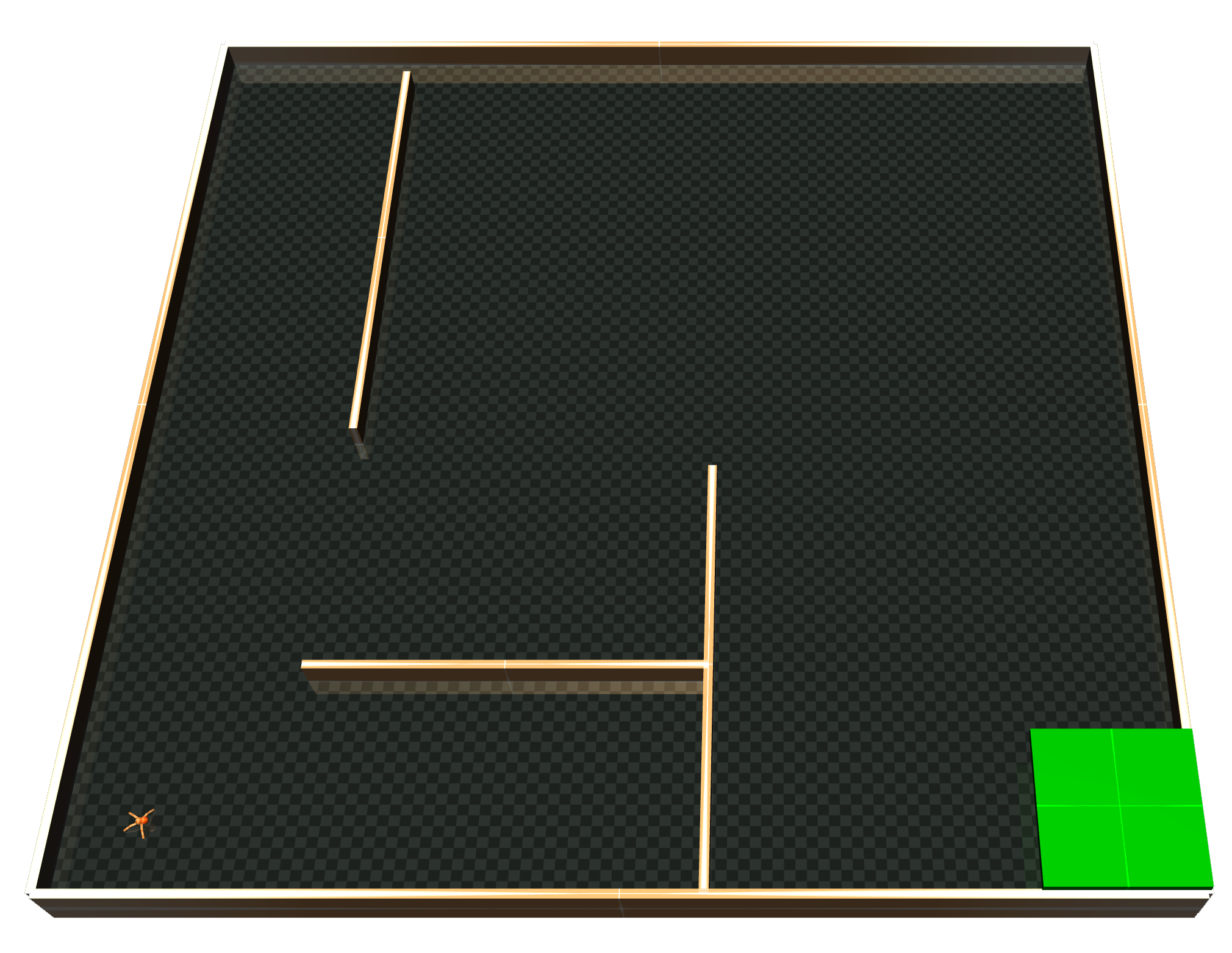}
   \caption{\antmaze}
   \label{fig:ant_maze}
 \end{subfigure}%
 \hfill
 \begin{subfigure}{.32\linewidth}
 \centering
   \includegraphics[width=1\linewidth]{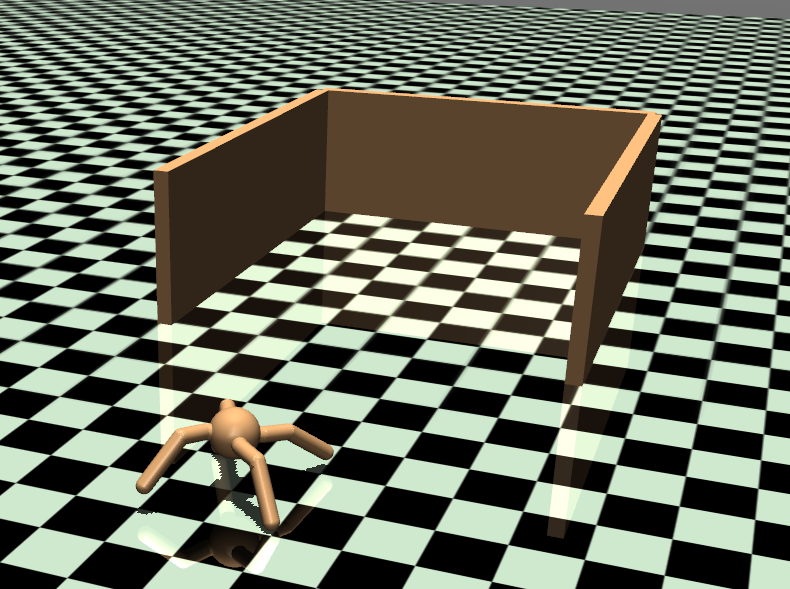}
   \caption{\anttrap}
   \label{fig:ant_trap}
 \end{subfigure}
     \caption{Evaluation environments. The state/action space in \ptmaze is $2 \times 2$, whereas they are $29 \times 8$ in \antmaze and $113 \times 8$ in \anttrap.}
     \label{fig:envs}
 \end{figure}
 
In the \ptmaze environment, an agent represented as a green sphere must find the exit of the maze depicted in \figurename~\ref{fig:point_maze_inertia}, represented as a red sphere. An observation contains the agent position at time $t$, and an action corresponds to position increments along the $x$ and $y$ axes. The reward is expressed as the negative Euclidean distance between the center of gravity of the agent and the exit center. The trajectory length cannot exceed 200 steps.

The \antmaze environment is modified from OpenAI Gym \ant \citep{brockman2016openai} and also used in \citep{colas2020scaling,frans2017meta}. In \antmaze, a four-legged ant starts in the bottom left of a maze and has to reach a goal zone located in the lower right part of it (green area in \figurename~\ref{fig:ant_maze}). As in \ptmaze, the reward is expressed as the negative Euclidean distance to the center of the goal zone. The final performance is the maximum reward received during an episode. The environment is considered solved when an agent obtains a score superior to $-10$. An episode consists of 3000 time steps, this horizon is three times larger than in usual \mujoco environments, making this environment particularly challenging for RL based methods \citep{vemula2019contrasting}.

Finally, the \anttrap environment also derives from \ant and is inspired from \citep{colas2020scaling, parker2020effective}. In \anttrap, the four-legged ant initially appears in front of a trap and must bypass it to run as fast as possible in the forward direction (see \figurename~\ref{fig:ant_trap}), as in \ant, the reward is computed as the ant velocity on the x-axis. The trap consists of three walls forming a dead-end directly in front of the ant. In this environment, the trajectory length cannot exceed 1000 steps. In the three environments, the state descriptor is defined as the agent center of gravity position at time step $t$ while the behavior descriptor is the center of gravity position at the end of the trajectory.
As opposed to \ptmaze and \antmaze, where the objective is to reach the exit area, there is no unique way to solve \anttrap and we expect a QD algorithm to generate various effective solutions as depicted in Figure~\ref{fig:teaser}. Furthermore, the behavior descriptor is not fully aligned with the fitness which makes it more difficult for \me and also for pure novelty seeking approaches as simply exploring the BD space is not enough to find a performing solution.

\subsection{Baselines and Ablations}

To answer question 1, we inspect the collection of solutions obtained with \qdpg and try to adapt to a new objective with a limited budget of 20 evaluations. \qdpg is then compared to three types of methods. To answer question 2, we compare \qdpg to a family of QD baselines, namely \mees, \nsres, \nsraes \citep{colas2020scaling}. Appendix~\ref{sec:supp_graphics} recaps the properties of all these methods. Then, to answer question 3, we compare \qdpg to a family of policy gradient baselines. Random Network Distillation (\rnd) \citep{burda2018exploration} and Adversarially Guided Actor Critic (\agac) \citep{AGAC2021} add curiosity-driven exploration process to a RL agent. We also compare to the population-based RL methods \cemrl, \pssstddd and to the closest method to ours, \pgame. Finally, to answer question 4, we propose to investigate the following matters: Can we replace alternating quality and diversity updates by a single update that optimizes for the sum of both criteria? Are quality (resp. diversity) gradients updates alone enough to fill the \me grid? Consequently, we consider the following ablations of \qdpg: \qdpgsum computes a gradient to optimize the sum of the quality and diversity rewards, \dpg applies only diversity gradients to the solutions, and \qpg applies only quality gradients, but both \dpg and \qpg still use QD selection. For all experiments, we limited the number of seeds to 5 due to limited compute capabilities.

%We compare \qdpg to its ablations and RL competitors in all environments and show results in Table~\ref{tab:comp_ablations_rl}. Like most practitioners, we only have access to limited resources, thus we could not re-evaluate \mees, \nsres and \nsraes on new environments as those methods are too computationally intensive (1000 CPUs for 2 days on \antmaze). We hence re-use the results from the original paper on \antmaze \citep{colas2020scaling}. Furthermore, as we decided to compare to many baselines, we had to limit the number of seeds to 5 for each one due to limited compute capabilities. More details are given in Appendix~\ref{sec:supp_results}.

\section{Results}
\label{sec:results}

\begin{figure*}[h!]
 \centering
 \begin{subfigure}{1\linewidth}
   \centering
   \includegraphics[width=1\linewidth]{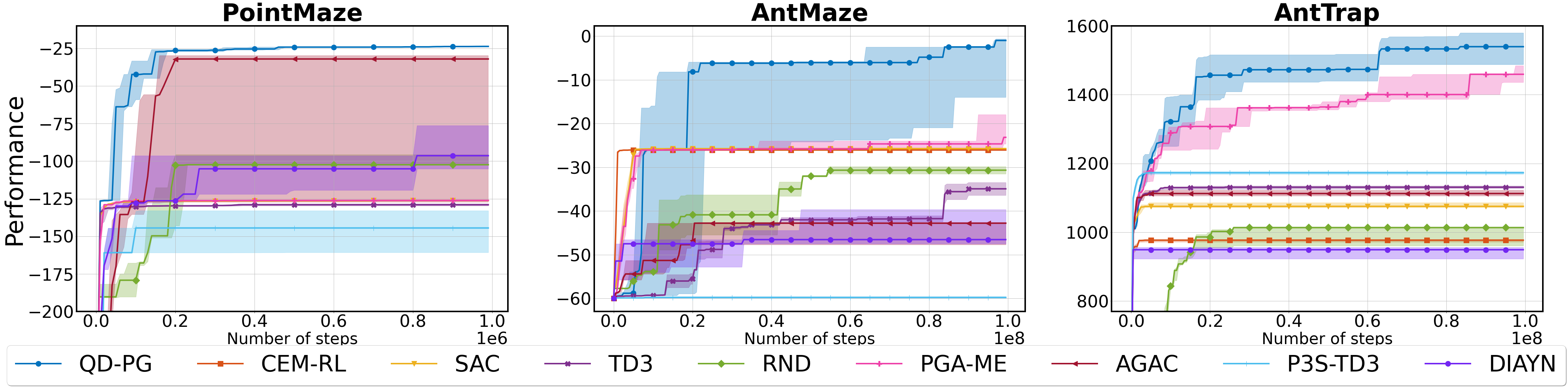}
   \caption{Performance of \qdpg and baseline algorithms on all environments. Plots present median bounded by first and third quartiles.}
   \label{fig:all_envs_comparison}
 \end{subfigure}%
 \newline
 \begin{subfigure}{1.\linewidth}
   \centering
   \hspace{2em}
   \includegraphics[width=0.2\linewidth]{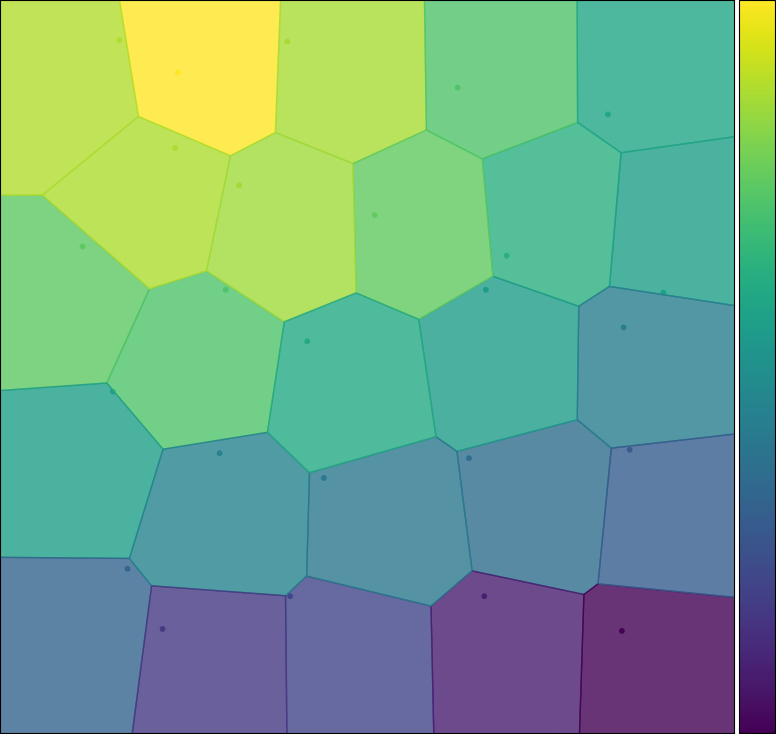}\hspace{7em}
   \includegraphics[width=0.2\linewidth]{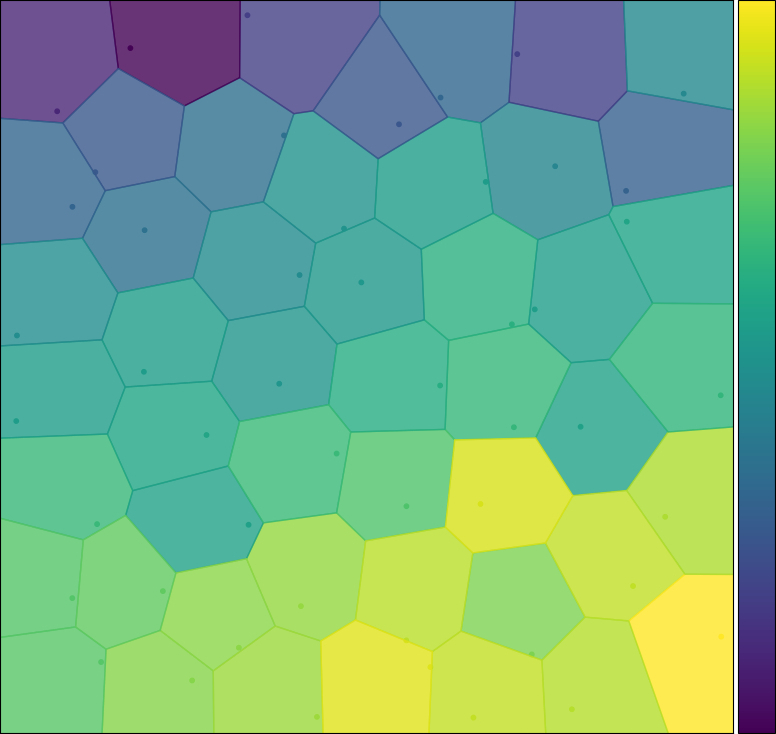}\hspace{5.5em}
   \raisebox{0.2\height}{\includegraphics[width=0.25\linewidth]{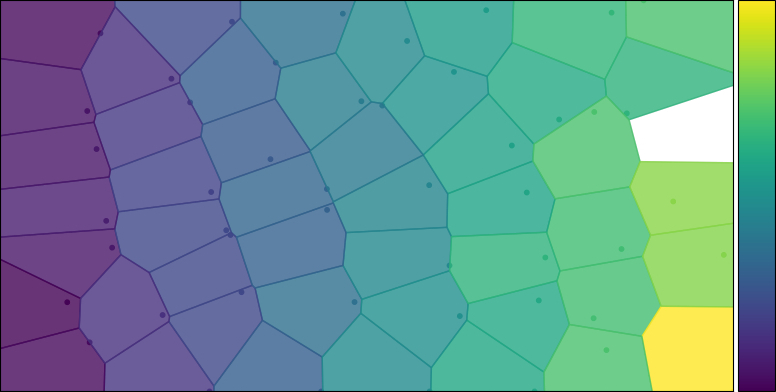}}
   \caption{\qdpg final grid on \ptmaze, \antmaze and \anttrap. Yellow and purple colors correspond respectively to high and low fitness.}
   \label{fig:ant_trap_grid}
 \end{subfigure}
     \caption{Performance of \qdpg and baseline algorithms and final grids of \qdpg for \ptmaze, \antmaze and \anttrap.}
     \label{fig:performance_results_comparison}
\end{figure*}

\textbf{Can QD-PG produce collections of neural policies and what are the advantages to do so?}\label{sec:fast_adaptation} Table~\ref{tab:comp_ablations_rl} and figure \ref{fig:performance_results_comparison} present \qdpg performances. In all environments, \qdpg manages to find working solutions that avoid local minima and reach the overall objective. In addition to its exploration capabilities, \qdpg generates collections of high performing solutions in a single run. During the \anttrap experiment, the final collection of solutions returned by \qdpg contained, among others, 5 solutions that were within a 10\% performance margin from the best one. As illustrated in Figure~\ref{fig:teaser}, these agents typically differ in their gaits and preferred trajectories to circumvent the trap.

\begin{figure}[h]
\centering
    \includegraphics[width=0.35\textwidth]{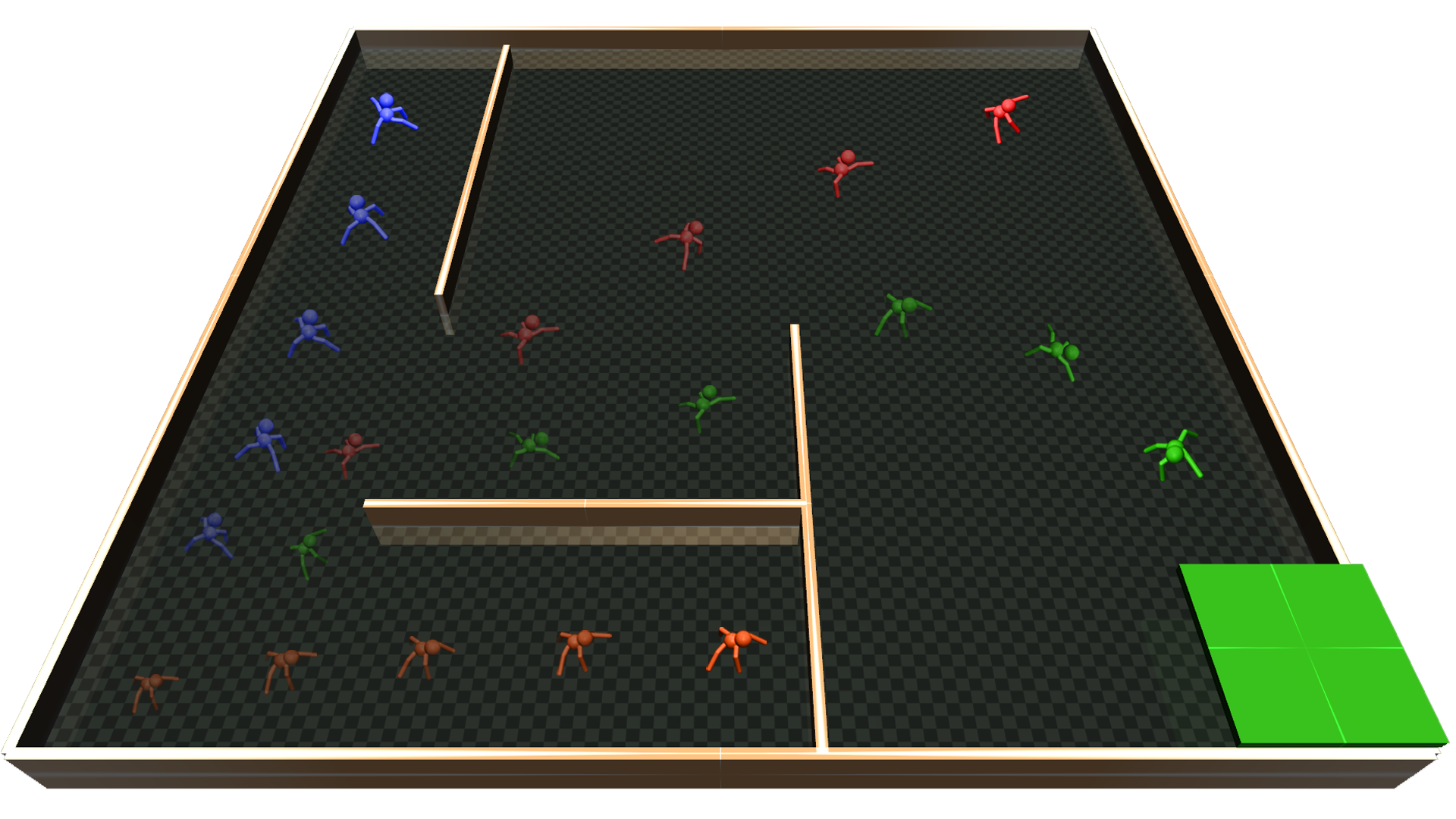}
    \caption{\qdpg produces a collection of diverse solutions. In \antmaze, even after setting new randomly located goals, the \me grid still contains solutions that are suited for the new objectives.}
    \label{fig:fast_adaptation}
\end{figure}

Generating a collection of diverse solutions comes with the benefit of having a repertoire of diverse solutions that can be used as alternatives when the MDP changes \citep{cully2015robots}.
We show that \qdpg is more robust than conventional policy gradient methods by changing the reward signal of the \antmaze environment. We replace the original goal in the bottom right part of the maze (see Figure~\ref{fig:fast_adaptation}) with a new randomly located goal in the maze.
Instead of running \qdpg to optimize for this new objective, we run a Bayesian optimization process to quickly find a good solution among the ones already stored in the grid.
With a budget of only $20$ solutions to be tested during the Bayesian optimization process, we are able to quickly recover a good solution for the new objective. We repeat this experiment $100$ times, each time with a different random goal, and obtain an average performance of $-10$ with a standard deviation of $9$.
In other words, $20$ interaction episodes (corresponding to $60.000$ time steps) suffice for the adaptation process to find a solution that performs well for the new objective without the need to re-train agents. Detailed results can be found in Appendix~\ref{sec:supp_fast_adaptation}.\\

\begin{table} %[!htb]
    \caption{Final performance compared to ablations and PG baselines. It is the minimum distance to the goal in \antmaze ($10^8$ steps) and the episode return in \ptmaze ($10^6$ steps) and \anttrap ($10^8$ steps). %Given numbers are medians and mean distances between median and first (resp. third) quartile.
    }
    %\begin{scriptsize}
    %\begin{subtable}[t]{\linewidth}
        \centering
        \label{tab:comp_ablations_rl}
        \begin{tabular}{cccl} %{l|ccc}
            \toprule
            %& \multicolumn{3}{c}{\textbf{Final Perf.} ($\pm$ std)}  \\
            %Final performance ($\pm$ std)\\
            %\midrule
            \textbf{Algorithm} & \ptmaze & \antmaze & \anttrap\\
            \midrule
            \qdpg & ${\bf -24 (\pm 0)}$ & ${\bf -1 (\pm 7)}$ & ${\bf 1540 (\pm 46)}$\\
            \qdpgsum & $-24 (\pm 0)$ & $-33 (\pm 4)$ & $1013 (\pm 0)$\\
            \dpg & $-36 (\pm 2)$ & $-59 (\pm 0)$ & $1014 (\pm 0)$\\
            \qpg & $-128 (\pm 0)$ & $-59 (\pm 0)$ & $1139 (\pm 38)$\\
            \midrule
            \sac & $-126 (\pm 0)$ & $-26 (\pm 0)$ & $1075 (\pm 7)$\\
            \tddd & $-129 (\pm 1)$ & $-35 (\pm 1)$ & $1131 (\pm 4)$\\
            \rnd & $-102 (\pm 4)$ & $-31 (\pm 1)$ & $1014 (\pm 27)$\\
            \midrule
            \cemrl & $-312 (\pm 1)$ & $-26 (\pm 0)$ & $977 (\pm 3)$\\
            \pssstddd & $-144 (\pm 14)$ & $-60 (\pm 0)$ & $1173 (\pm 4)$\\
            \agac & $-32 (\pm 49)$ & $-43 (\pm 3)$ & $1113 (\pm 8)$\\
            \diayn & $-96 (\pm 14)$ & $-47 (\pm 4)$ & $949 (\pm 34)$\\
            \pgame & $-126 (\pm 0)$ & $-18 (\pm 6)$ & $1455 (\pm 17)$\\
            \bottomrule
            %\caption{Comparison to ablations and PG baselines. }
        \end{tabular}
    %\end{subtable}%
\end{table}

\textbf{Is \qdpg more sample efficient than its evolutionary competitors?} Table~\ref{tab:comp_evo} compares \qdpg to Deep Neuroevolution algorithms with a diversity seeking component (\mees, \nsres, \nsraes, \cemrl) in terms of sample efficiency. \qdpg runs on 10 CPU cores for 2 days while its competitors used 1000 \cpu cores for the same duration. Nonetheless, \qdpg matches the asymptotic performance of \mees using two orders of magnitude fewer samples, explaining the lower resource requirements.

We see three reasons for the improved sample efficiency of \qdpg: 1) \qdpg leverages a replay buffer and can re-use each sample several times. 2) \qdpg leverages novelty at the state level and can exploit all collected transitions to maximize quality and diversity. For instance, in \antmaze, a trajectory brings 3000 samples to \qdpg while standard QD methods would consider it a unique sample. 3) PG exploits the analytical gradient between the neural network weights and the resulting policy action distribution and estimates only the impact of the distribution on the return. By contrast, standard QD methods directly estimate the impact on the return of randomly modifying the weights.\\

\begin{table}
    %\begin{subtable}[t]{\linewidth}
        \centering
        \caption{Comparison to evolutionary competitors on \antmaze. The {\bf Ratio} column compares the sample efficiency of a method to \qdpg.}
        \label{tab:comp_evo}
        \begin{tabular}{cccl} %{l|ccc}
        % & \multicolumn{3}{c}{\antmaze} \\
        \toprule
        \textbf{Algorithm} & \textbf{Final Perf.} & \textbf{Steps to goal} & \textbf{Ratio}\\
        \midrule
        \qdpg & $-1 (\pm 7)$ & ${\bf 8.4e7}$ & {\bf 1}\\
        \cemrl & $-26 (\pm 0)$ & $\infty$ & $\infty$\\
        \mees & $-5 (\pm 0)$ & $2.4e10$ & 286\\
        \nsres & $-26 (\pm 0)$ & $\infty$ & $\infty$\\
        \nsraes & ${\bf -2 (\pm 1)}$ & $2.1 e10$ & 249\\
        \bottomrule
        \end{tabular}
    %\end{subtable}
    %\end{scriptsize}
\end{table}

\textbf{How challenging are the considered benchmarks for policy gradients methods?} Table~\ref{tab:comp_ablations_rl} compares \qdpg to state-of-the-art policy gradient algorithms and validates that classical policy gradient methods fail to find optimal solutions in deceptive environments. \tddd quickly converges to local minima of performance resulting from being attracted in dead-ends by the deceptive gradients. While we may expect \sac to better explore due to entropy regularization, it also converges to that same local minima.
 
Besides, despite its exploration mechanism based on \cem, \cemrl also quickly converges to local optima in all benchmarks, confirming the need for a dedicated diversity seeking component.
\rnd, which adds an exploration bonus used as an intrinsic reward, also demonstrates performances inferior to \qdpg in all environments but manages to solve \ptmaze. In \anttrap, \rnd extensively explores the BD space but fails to obtain high returns. Although maintaining diversity into a population of reinforcement learning agents, \agac and \pssstddd are not able to explore enough of the environments to solve them, showing some limits of exploration through the action space.

\diayn is able to explore in \ptmaze but does not fully explore the environment and hence does not reach the goal before the end of the $10^6$ steps. \diayn ensures that its skills explore states different enough to be discriminated, but once they are, there is no incentive to further explore. Moreover, \diayn in \anttrap shows a limit of unsupervised learning methods: \anttrap is an example of environment where the behavior descriptor chosen is not perfectly aligned with the performance measured. As rewards are completely ignored by \diayn, the produced controllers have very low fitness.\\

Figure \ref{fig:performance_results_comparison} relates performance of \qdpg and \pgame along time on \ptmaze, \antmaze, \anttrap and shows that \qdpg outperforms the latter both in final performance and data-efficiency by a significant margin. It was the first time that \pgame was assessed in exploration environments. This definitely confirms that the Genetic mutation used by \pgame struggles in big parameter space while the Diversity Policy Gradient offers a data-efficient diversity-seeking mechanism to improve the exploration of \me in high-dimensional deceptive environments. 

We tried \qdpg on the locomotion tasks from the original paper \citep{Nilsson2021} and found it hard to design state descriptor extractors $\Psi$ such that our diversity rewards would be strongly correlated with the behavior descriptor $\xi$ used in \pgame \citep{Nilsson2021} in the sense of equation \ref{eq:diversity_gradient_constraint}, showing a limitation of our work. This leads us to the conclusion that keeping genetic mutations in addition to our diversity policy gradient could end up in a fruitful and robust synergy. We leave this for future work. Finally, we observed that the learning process of \qdpg on \antmaze sometimes struggled to start, explaining the high interquartile distance of performances on \antmaze (figure \ref{fig:performance_results_comparison}).\\

\textbf{What is the usefulness of the different components of \qdpg ?} The ablation study in \tableautorefname{}~\ref{tab:comp_ablations_rl} shows that when maximising quality only, \qpg fails due to the deceptive nature of the reward and when maximizing diversity only, \dpg sufficiently explores to solve \ptmaze but requires more steps and finds lower-performing solutions. When optimizing simultaneously for quality and diversity, \qdpgsum fails to learn in \anttrap and manages to solve the task in \antmaze but requires more samples than \qdpg. We hypothesize that quality and diversity rewards may give rise to conflicting gradients. Therefore, both rewards cancel each other, preventing any learning.
This study validates the usefulness of \qdpg components: 1) optimizing for diversity is required to overcome the deceptive nature of the reward; 2) adding quality optimization provides better asymptotic performance; 3) it is better to disentangle quality and diversity updates.

\section{Conclusion}

This paper is the first to introduce a diversity gradient to seek diversity both at the state and episode levels. Based on this component, we proposed a novel algorithm, \qdpg, that builds upon MAP-Elites algorithm and uses methods inspired from reinforcement learning literature to produce collections of diverse and high-performing neural policies in a sample-efficient manner. We showed experimentally that \qdpg generates several solutions that achieve high returns in challenging exploration problems. Although more efficient than genetic mutations, our diversity policy gradient requires a state descriptor that will ensure a correlation between state diversity and global diversity, which can be difficult when the behavior descriptor is more complex. We believe that an interesting mitigation would be to combine our diversity seeking component with a genetic mutation \citep{Vassiliades2018} to have robust exploration in a wider range of situations.


\begin{acks}

We would like to thank Claire Bizon Monroc and Raphael Boige for their help in reproducing some baselines and for their helpful comments. Work by Nicolas Perrin-Gilbert was partially supported by the French National Research Agency (ANR), Project ANR-18-CE33-0005 HUSKI.

\end{acks}

\bibliographystyle{ACM-Reference-Format}
\bibliography{qdpg}

\clearpage
\newpage

\appendix

\section{QD-PG Code and architecture}
\label{sec:algo_supp}

\begin{figure*}[thbp!]
\centering
\begin{subfigure}{.17\linewidth}
\centering
  \includegraphics[width=1\linewidth]{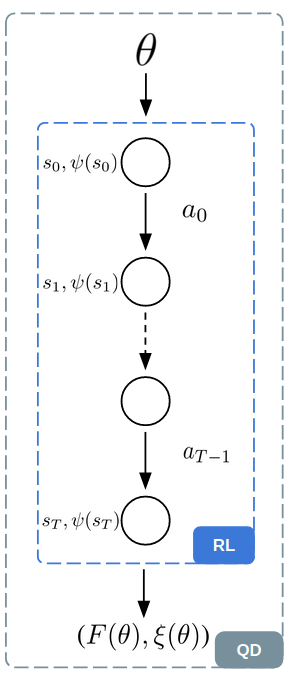}
  \caption{}
  \label{fig:two_levels}
\end{subfigure}%
\hfill
\begin{subfigure}{.83\linewidth}
\centering
  \includegraphics[width=1\linewidth]{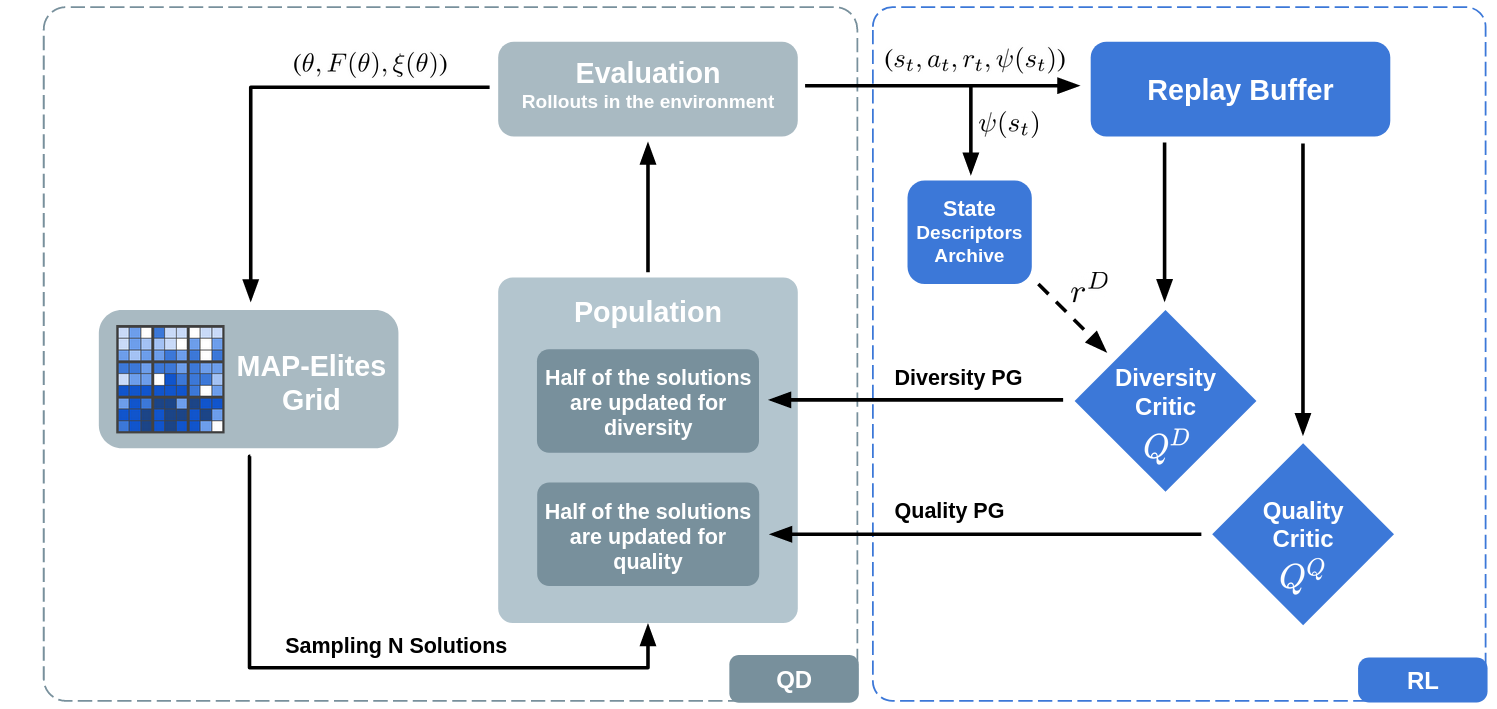}
  \caption{}
  \label{fig:general_schema}
\end{subfigure}
    \caption{(a): The RL part of \qdpg operates at the time step level while the QD part operates at the controller level, considering the MDP as a black box. (b) One \qdpg iteration consists of three phases: 1) A new population of solutions is sampled from the \me grid. 2) These solutions are updated by an off-policy RL agent: half of the solutions are optimized for quality and the other half for diversity. The RL agent leverages one shared critic for each objective. 3) The newly obtained solutions are evaluated in the environment. Transitions are stored in a replay buffer while the updated solutions, their final scores and behavior descriptors are stored in the \me grid.\label{fig:archi}}
\end{figure*}

The full algorithm can be found on pseudocode \ref{alg:full_pseudocode} and its architecture in Figure \ref{fig:archi}.
\makeatletter
\newcommand{\removelatexerror}{\let\@latex@error\@gobble}
\makeatother

\begin{figure*}[h!]
\removelatexerror
\centering
\resizebox{0.8\textwidth}{!}{%
\begin{algorithm}[H]
    \small
    \SetAlgoLined
    \DontPrintSemicolon
    \SetKwInput{KwInput}{Given}
    \KwInput{N, max\_steps, gradient\_steps\_ratio, BD extraction function $\xi$, state descriptor extraction function $\psi$}
    \SetKwInput{KwInput}{Initialize}
    \KwInput{MAP-Elites grid $\mathbb{M}$, Replay Buffer $\mathbb{R}$, $N$ actors $\{\pi_{\theta_{i}}\}_{i=\{1,\dots,N\}}$, 2 critics $Q^D_w$ and $Q^Q_w$, state descriptors archive $\mathbb{A}$}\;

    $total\_steps, actor\_steps = 0, 0 $ \texttt{// Step counters}\;
    \texttt{\\}
    
    \tcp{Parallel evaluation of the initial population}
    \For{$j \leftarrow 1$ \KwTo $N$}{ 
        Play one episode with actor $\pi_{\theta_{j}}$\ and store all transitions in $\mathbb{R}$\;
        Get episode length $T$, discounted return $R$ and state descriptors $\{\psi(s_1), \dots, \psi(s_T)\}$\;
        Store state descriptors $\{\psi(s_1), \dots, \psi(s_T)\}$ in $\mathbb{A}$\;
        Compute $\xi(\theta_j)$ and add the tuple ($R$, $\xi(\theta_j)$, $\theta_j$) in the MAP-Elites grid $\mathbb{M}$\;
        $actor\_steps \leftarrow actor\_steps + T$\;
    }
    \texttt{\\}
    
    \tcp{Main loop}
    \While{$total\_steps<max\_steps$}{
    \texttt{\\}
    
        \tcp{Select new generation}
        %Compute the Pareto Front form the archive $A$\;
        Get $N$ actors $\pi_{\theta_i}, i \in \{1, \dots, N\}$ from $\mathbb{M}$\;
        $gradient\_steps$ = $int(actor\_steps \times gradient\_steps\_ratio)$\;
        $actor\_steps=0$\;
        \texttt{\\}
        
        \tcp{Perform in parallel population update and evaluation}
        \For{$j \leftarrow 1$ \KwTo N}{
            \texttt{\\}
            \tcp{Update the population}
            \For{$i \leftarrow 1$ \KwTo $gradient\_steps$}{
                \texttt{\\}
                Sample batch of $(s_t, a_t, r_t, s_{t+1}, \psi(s_t))$ from $\mathbb{R}$\;
                \texttt{\\}
                \tcp{First half is updated to maximise diversity}
                \If{$j \leq $N$ // 2$}{
                    Compute novelty reward as $r^D_t$ from $\psi(s_t)$ and $\mathbb{A}$\;
                    Update $\pi_{\theta_j}$ for diversity\;
                    Compute the novelty critic gradient locally \;
                    Average novelty critic gradients between threads\;
                    Update novelty critic $Q^D_{w}$\;
                }
                \texttt{\\}
                \tcp{Second half is updated to maximise quality}
                \Else{
                Update $\pi_{\theta_j}$ for quality \;
                    Compute the quality critic gradient locally \;
                    Average quality critic gradients between threads \;
                    Update quality critic $Q^Q_{v}$ \;
                }
            }
            \texttt{\\}
            
            \tcp{Evaluate the updated actors}
        Play one episode with actor $\pi_{\theta_{j}}$\ and store all transitions in $\mathbb{R}$\;
        Get episode length $T$, discounted return $R$ and state descriptors $\{\psi(s_1), \dots, \psi(s_T)\}$\;
        Store state descriptors $\{\psi(s_1), \dots, \psi(s_T)\}$ in $\mathbb{A}$\;
        Compute $\xi(\theta_j)$ and add the tuple ($R$, $\xi(\theta_j)$, $\theta_j$) in the MAP-Elites grid $\mathbb{M}$\;
        $actor\_steps \leftarrow actor\_steps + T$\;
        }
        \texttt{\\}
        
        $total\_steps \leftarrow total\_steps + actor\_steps$ \tcp{Update total time steps}\;
    }
    
    \caption{QD-PG}
    \label{alg:full_pseudocode}
\end{algorithm}
}
\end{figure*}

\section{Detailed Equations of TD3}
\label{sec:td3}
The Twin Delayed Deep Deterministic (\tddd) agent \cite{fujimoto2018addressing} builds upon the Deep Deterministic Policy Gradient (\ddpg) agent \cite{lillicrap2015continuous}. It trains a deterministic actor $\pi_{\phi}: \mathcal{S} \rightarrow \mathcal{A}$ directly mapping observations to continuous actions and a critic $Q_{\theta}: \mathcal{S}\times \mathcal{A} \rightarrow \reals$ which takes a state $s$ and an action $a$ and estimates the average return from selecting action $a$ in state $s$ and then following policy $\pi_{\phi}$. As \ddpg, \tddd alternates between policy evaluation and policy improvement so as to maximise the average discounted return. In \ddpg, the critic is updated to minimize a temporal difference error during the policy evaluation step which induces an overestimation bias. \tddd corrects for this bias by introducing two critics $Q_{\theta_1}$ and $Q_{\theta_2}$. \tddd plays one step in the environment using its deterministic policy and then stores the observed transition $(s_t, a_t, r_t, s_{t+1})$ into a replay buffer $\mathcal{M}$. Then, it samples a batch of transitions from $\mathcal{M}$ and updates the critic networks. Half the time it also samples another batch of transitions to update the actor network.

Both critics are updated so as to minimize a loss function which is expressed as a mean squared error between their predictions and a target:

\begin{equation}
\label{eq:td3_critic}
    L^{critic}(\theta_1 , \theta_2) = \sum\limits_{\text{batch}}\sum\limits_{i=1,2} (Q_{\theta_i}(s_t, a_t) - y_t)^2,
\end{equation}

where the common target $y_t$ is computed as:

\begin{equation}
    y_t = r_t + \gamma \min\limits_{i=1,2} Q_{\theta_i}(s_{t+1}, \pi_{\phi}(s_{t+1}) + \epsilon), 
\end{equation}

where $\epsilon \sim \mathcal{N}(0,I)$.

The Q-value estimation used to compute target $y_t$ is taken as minimum between both critic predictions thus reducing the overestimation bias. \tddd also adds a small perturbation $\epsilon$ to the action $\pi_{\phi}(s_{t+1})$ so as to smooth the value estimate by bootstrapping similar state-action value estimates.

Every two critics updates, the actor $\pi_{\phi}$ is updated using the deterministic policy gradient also used in \ddpg \cite{silver2014deterministic}. For a state $s$, \ddpg updates the actor so as to maximise the critic estimation for this state $s$ and the action $a=\pi_{\phi}(s)$ selected by the actor. As there are two critics in \tddd, the authors suggest to take the first critic as an arbitrary choice. The actor is updated by minimizing the following loss function:

\begin{equation}
\label{eq:td3_actor}
    L^{actor}(\phi) = - \sum\limits_{\text{batch}} Q_{\theta_1}(s_t, \pi_{\phi}(s_t)).
\end{equation}

Policy evaluation and policy improvement steps are repeated until convergence. \tddd demonstrates state of the art performance on several \mujoco benchmarks.

\section{QD-PG  Details}
\label{sec:qdrl_details}

\subsection{Computational details}
We consider populations of $N=4$ actors for the \ptmaze environment and $N=10$ actors for \antmaze and \anttrap. We use 1 \cpu thread per actor and parallelization is implemented with the Message Passing Interface (MPI) library. Our experiments are run on a standard computer with $10$ \cpu cores and 100 {\sc GB} of {\sc RAM}, although the maximum RAM consumption per experiment at any time never exceeds 10GB due to an efficient and centralized management of the \me grid which stores all solutions. An experiment on \ptmaze typically takes between 2 and 3 hours while an experiment on \antmaze or \anttrap takes about 2 days. Note that these durations can vary significantly depending on the type of \cpu used. We did not use any \gpu.

Computational costs of \qdpg mainly come from backpropagation during the update of each agent, and to the interaction between agents and the environment. These costs scale linearly with the population size but, as many other population-based methods, the structure of \qdpg lends itself very well to parallelization. We leverage this property and parallelize our implementation to assign one agent per \cpu thread. Memory consumption also scales linearly with the number of agents. To reduce this consumption, we centralize the \me grid on a master worker and distribute data among workers when needed. With these implementation choices, \qdpg only needs a very accessible computational budget for all experiments.

\subsection{MAP-Elites Implementation details}
\qdpg uses a \me grid as archive of solutions. We assume that the BD space is bounded and can be discretized into an Cartesian grid. We discretize each dimension into $m$ meshes, see \tableautorefname{}~\ref{tab:hyperparameters} for the value of $m$ depending on the environment. Hence, the number of cells in the \me grid equals $m$ times the number of dimensions of the BD space.  When a new solution $\theta$ is obtained after the mutation phase, we look for the cell corresponding to its BD, $\xi(\theta)$. If the cell is empty, the solution is added, otherwise the new solution replaces the solution already contained in the cell if its score $F(\theta)$ is better than the score of the already contained solution. During selection, we sample solutions uniformly from the \me grid.

\subsection{Diversity reward computation}
\qdpg optimizes solutions for quality but also for diversity at the state level. The diversity policy gradient updates the solutions so as to encourage them to visit states with novel state descriptors. The novelty of a state descriptor $\psi(s_t)$ is expressed through a diversity reward $r_t^D$. In practice, we maintain a FIFO archive $\mathbb{A}$ of the state descriptors encountered so far. When a transition $(s_t, a_t, r_t, s_{t+1}, \psi(s_t))$ is stored in the replay buffer, we also add $\psi(s_t)$ to $\mathbb{A}$. We only add a state descriptor in $\mathbb{A}$ if its mean Euclidean distance to its $K$ nearest neighbors is greater than an acceptance threshold. This filtering step enables to keep the archive size  reasonable and to facilitate the computation of the $K$ nearest neighbors. The values of $K$ and of the threshold are given in Table~\ref{tab:hyperparameters}. When a batch of transitions is collected during the update phase, we recompute fresh diversity rewards $r_t^D$ as the mean Euclidean distance between the sampled state descriptors $\psi(s_t)$ and their $K$ nearest neighbors in $\mathbb{A}$. These diversity rewards are used instead of standard rewards in sampled transitions $(s_t, a_t, r_t^D, s_{t+1}, \psi(s_t))$ to compute the diversity policy gradient.

\subsection{QD-PG Hyper-parameters}
\label{sec:qdrl_hp}
\tableautorefname{}~\ref{tab:hyperparameters} summarizes hyper-parameters used in experiments. Most of these hyper-parameter values are taken from \tddd.

\begin{table}[thbp!]
    \centering
    \caption{\qdpg Hyper-parameters: \antmaze and \anttrap hyper-parameters are identical and grouped under the Ant column}
    \label{tab:hyperparameters}
    \begin{tabular}{lll}
        \toprule
        \textbf{Parameter} & \textbf{PointMaze} & \textbf{Ant}\\
        \midrule
        \\
        \multicolumn{3}{l}{\textbf{TD3}}\\
        \midrule
        Optimizer & SGD & SGD \\
        Learning rate & $6.10^{-3}$ & $3.10^{-4}$ \\
        Discount factor $\gamma$ & $0.99$ & $0.99$ \\
        Replay buffer size & $10^{6}$ & $5.10^{5}$ \\
        Hidden layers size & $64/32$ & $256/256$\\
        Activations & ReLU & ReLU \\
        Minibatch size & $256$ & $256$ \\
        Target smoothing coeff. & $0.005$ & $0.005$ \\
        Delay policy update & $2$ & $2$\\
        Target update interval & $1$ & $1$\\
        Gradient steps ratio & $4$ & $0.1$\\
        \\
        \multicolumn{3}{l}{\textbf{State Descriptors Archive}}\\
        \midrule
        Archive size &  $10000$ & $10000$ \\
        Threshold of acceptance & $0.0001$ & $0.1$\\
        K-nearest neighbors & $10$ & $10$ \\
        \\
        \multicolumn{3}{l}{\textbf{MAP-Elites}}\\
        \midrule
        Nb. of bins per dimension &  $5$ & $7$\\
        \bottomrule
  \end{tabular}
\end{table}

\section{Environments analysis}
\label{sec:analysis_supp}

In \ptmaze, the state and action spaces are two-dimensional. By contrast, in \antmaze and \anttrap, the dimensions of their observation spaces are respectively equal to 29 and 113 while the dimensions of their action spaces are both equal to 8, making these two environments much more challenging as they require larger controllers.
The \anttrap environment also differs from mazes as it is open-ended, i.e., the space to be explored by the agent is unlimited, unlike mazes where this space is restricted by the walls. In this case, a state descriptor corresponds to the ant position that is clipped to remain in a given range. On the $y$-axis, this range is defined as three times the width of the trap. On the $x$-axis, this range begins slightly behind the starting position of the ant and is large enough to let it accelerate along this axis. \figurename~\ref{fig:coverage_map_anttrap} depicts the BD space in \anttrap.

In all environments, state descriptors $\psi(s_t)$ are defined as the agent's position at time step $t$ and behavior descriptors $\xi(\theta)$ are defined as the agent's position at the end of a trajectory. Therefore, we have $\mathcal{B} = \mathcal{D} = \mathbb{R}^2$, $\psi(s_t) = (x_t, y_t)$ and $\xi(\theta) = (x_T, y_T)$ where $T$ is the trajectory length. We also take $||.||_{\mathcal{B}}$ and $||.||_{\mathcal{D}}$ as Euclidean distances. This choice does not always satisfy Equation~\eqref{eq:diversity_gradient_constraint} but is convenient in practice and led to satisfactory results. The peculiarity of \anttrap lies in the fact that the reward is expressed as the forward velocity of the ant, thus making the descriptors not totally aligned with the task.

\begin{figure}[ht!]
\centering
\begin{subfigure}{.32\linewidth}
  \includegraphics[width=1\linewidth]{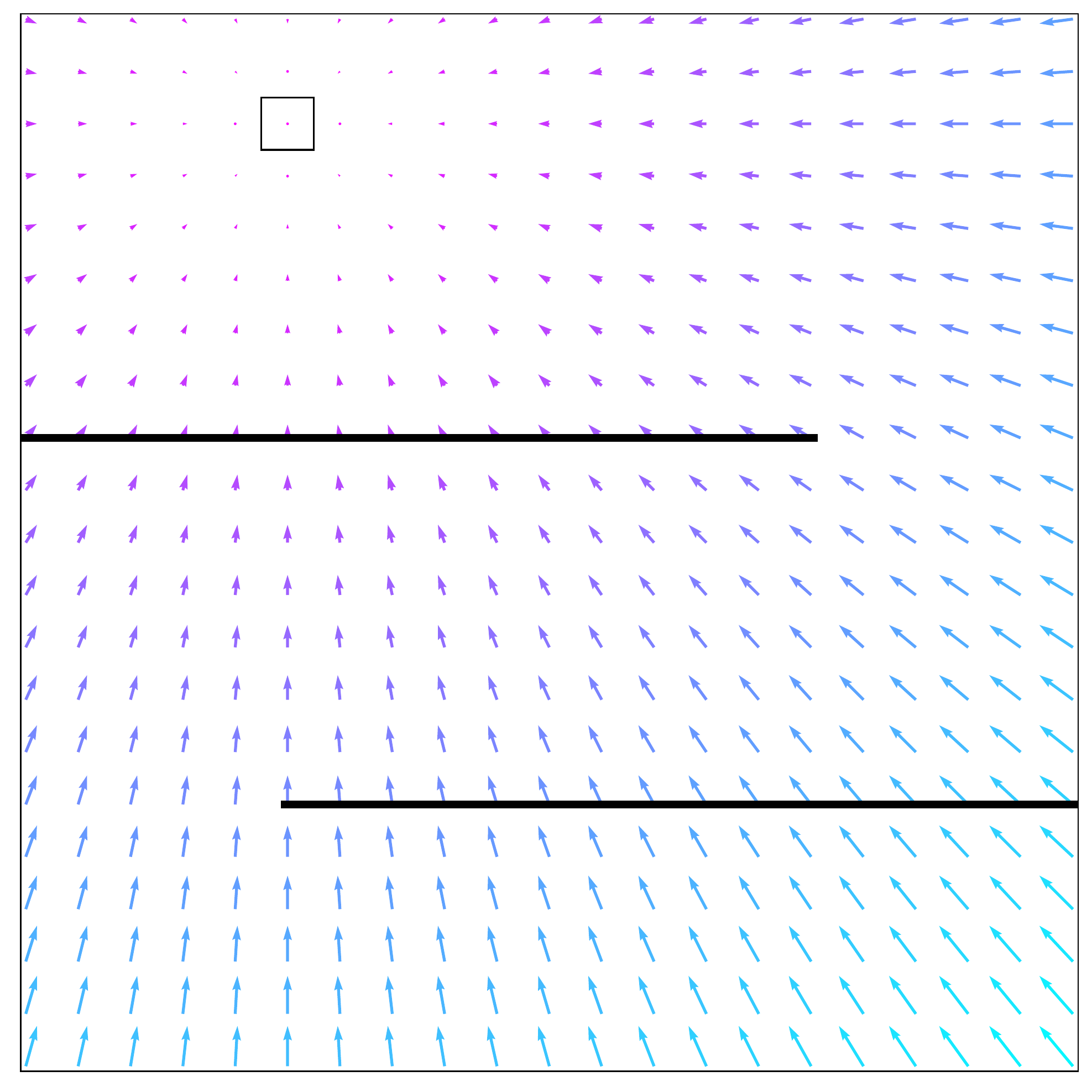}
  \caption{\ptmaze}
  \label{fig:gmap_ptmaze}
\end{subfigure}
\hspace{0.1\linewidth}
\begin{subfigure}{.33\linewidth}
  \includegraphics[width=1\linewidth]{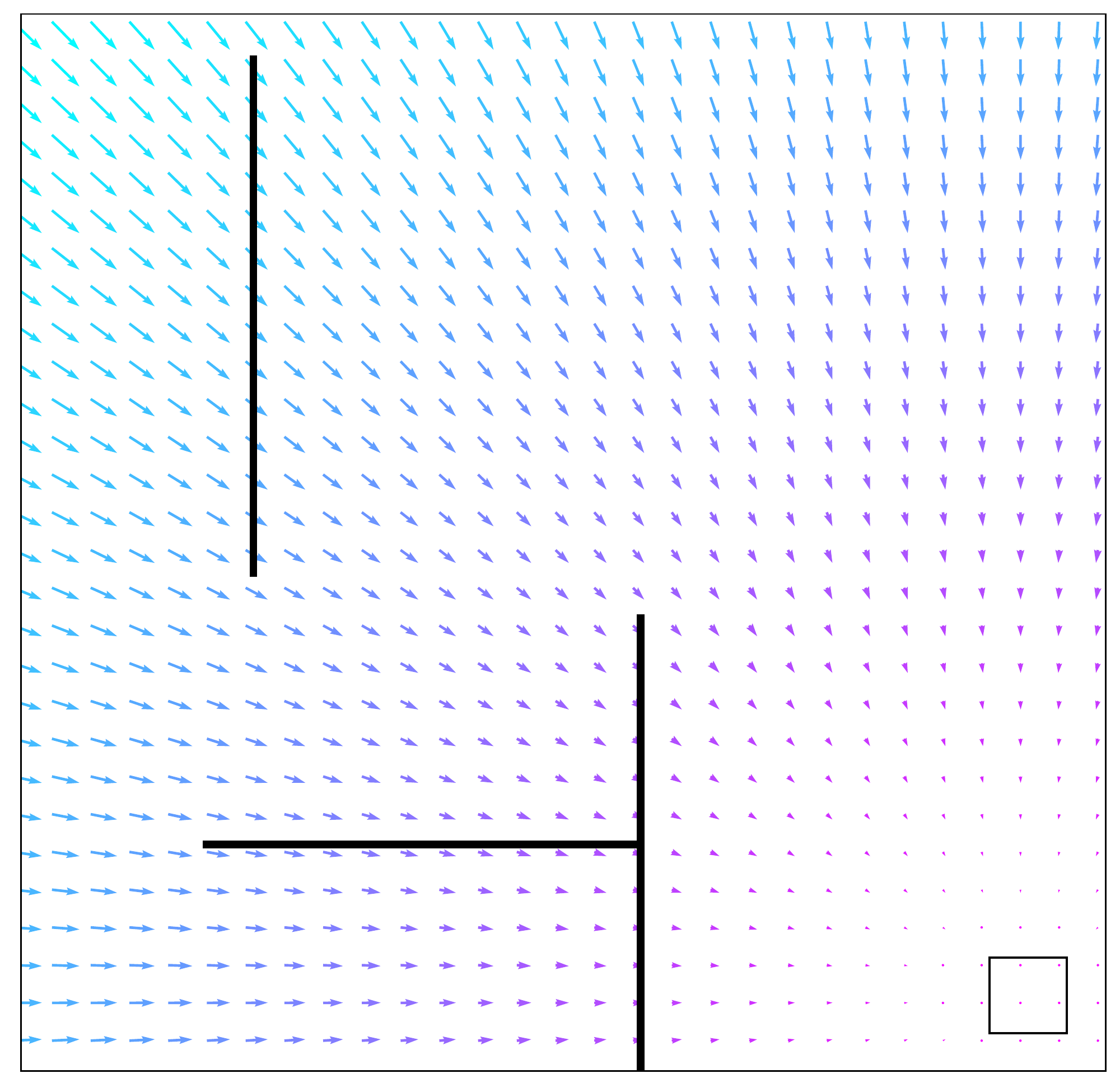}
  \caption{\antmaze}
  \label{fig:gmap_antmaze}
\end{subfigure}
    \caption{Gradients maps on \ptmaze and \antmaze. Black lines are maze walls, arrows depict gradient fields and the square indicates the maze exit. Both settings present deceptive gradients as naively following them leads into a wall.}
    \label{fig:grad_maps}
\end{figure}

\figureautorefname{}~\ref{fig:grad_maps} highlights the deceptive nature of the \ptmaze and the \antmaze objective functions by depicting gradient fields in both environments. Similarly, the reward is  also deceptive in \anttrap.

\section{Detailed results}
\label{sec:supp_results}
In this section, we provide performance charts corresponding to \tableautorefname{}~\ref{tab:comp_evo} and \tableautorefname{}~\ref{tab:comp_ablations_rl}, coverage maps highlighting the exploration capabilities of \qdpg, and detailed results of the fast adaptation experiment. \tableautorefname{}~\ref{tab:recap} summarizes the different components present in \qdpg, its ablations and all baselines.

\subsection{Performance charts}
\label{sec:supp_graphics}
\begin{figure}[thbp!]
\centering
\begin{subfigure}{.5\linewidth}
\centering
  \includegraphics[width=1\linewidth]{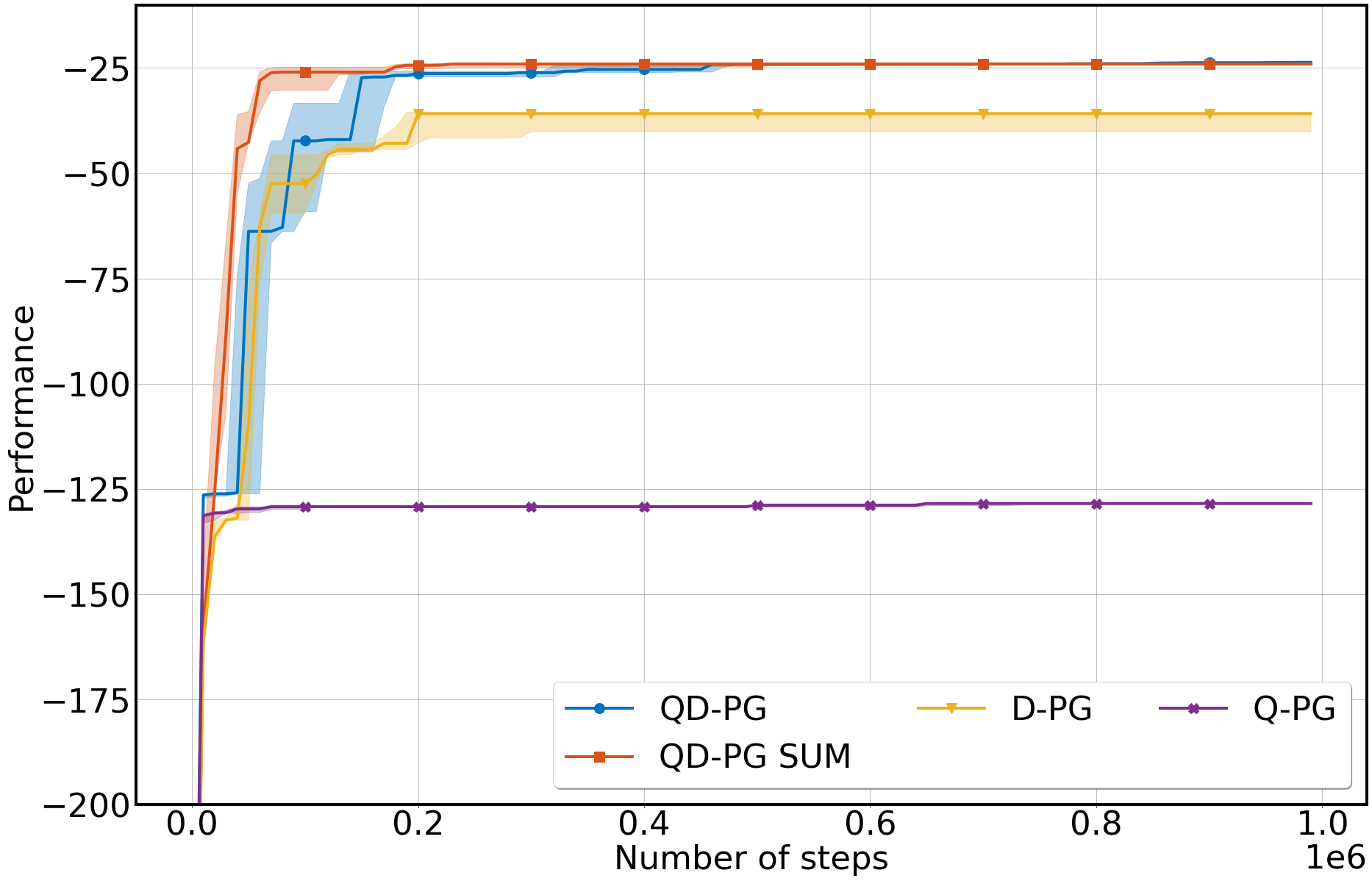}
  \caption{\ptmaze: Ablation study}
  \label{fig:ptmazeinertia_ablation}
\end{subfigure}%
\hfill
\begin{subfigure}{.5\linewidth}
\centering
  \includegraphics[width=1\linewidth]{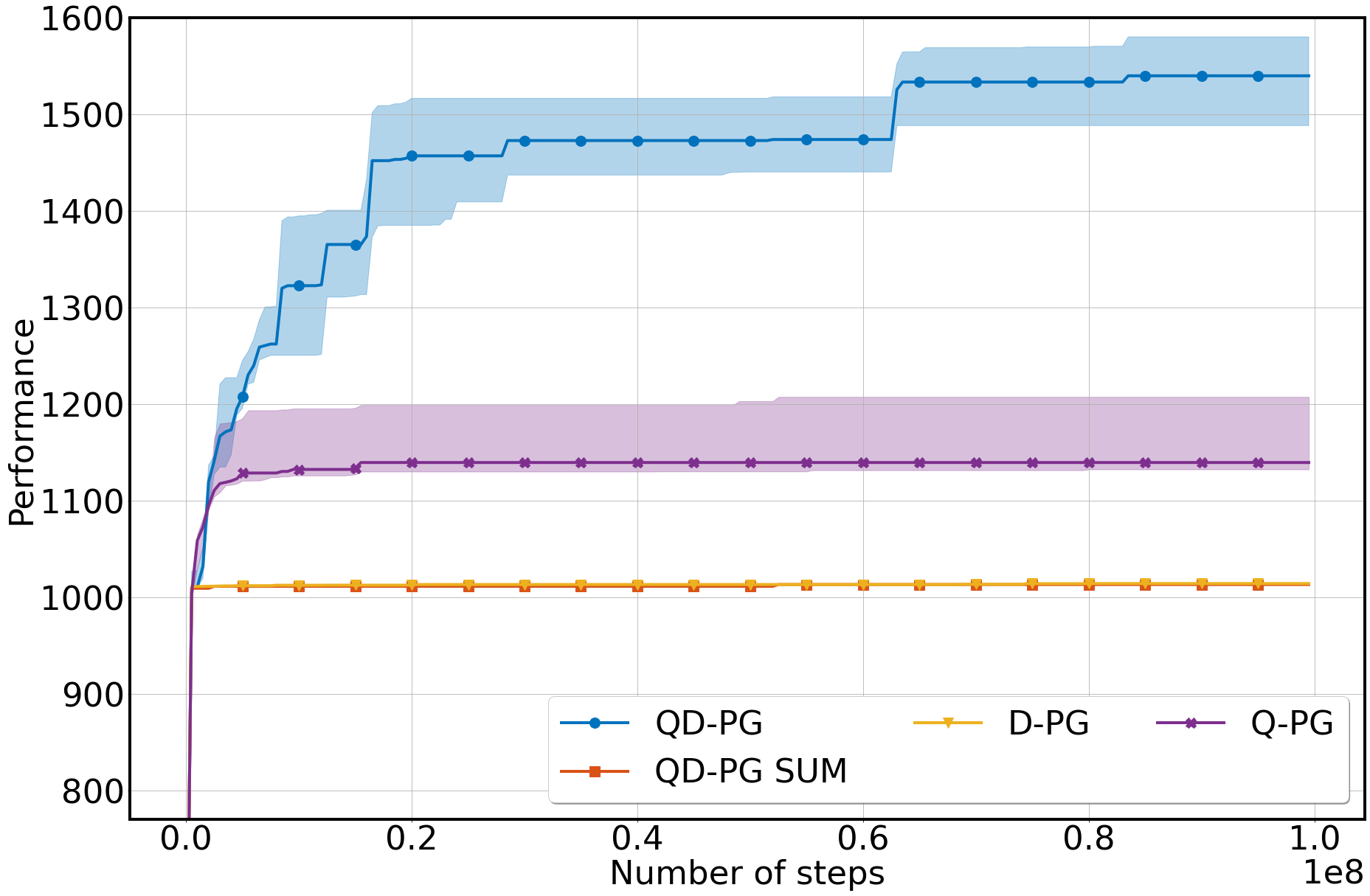}
  \caption{\anttrap: Ablation study}
  \label{fig:qdrlanttrap_ablation}
\end{subfigure}
\hfill
\begin{subfigure}{.5\linewidth}
\centering
  \includegraphics[width=1\linewidth]{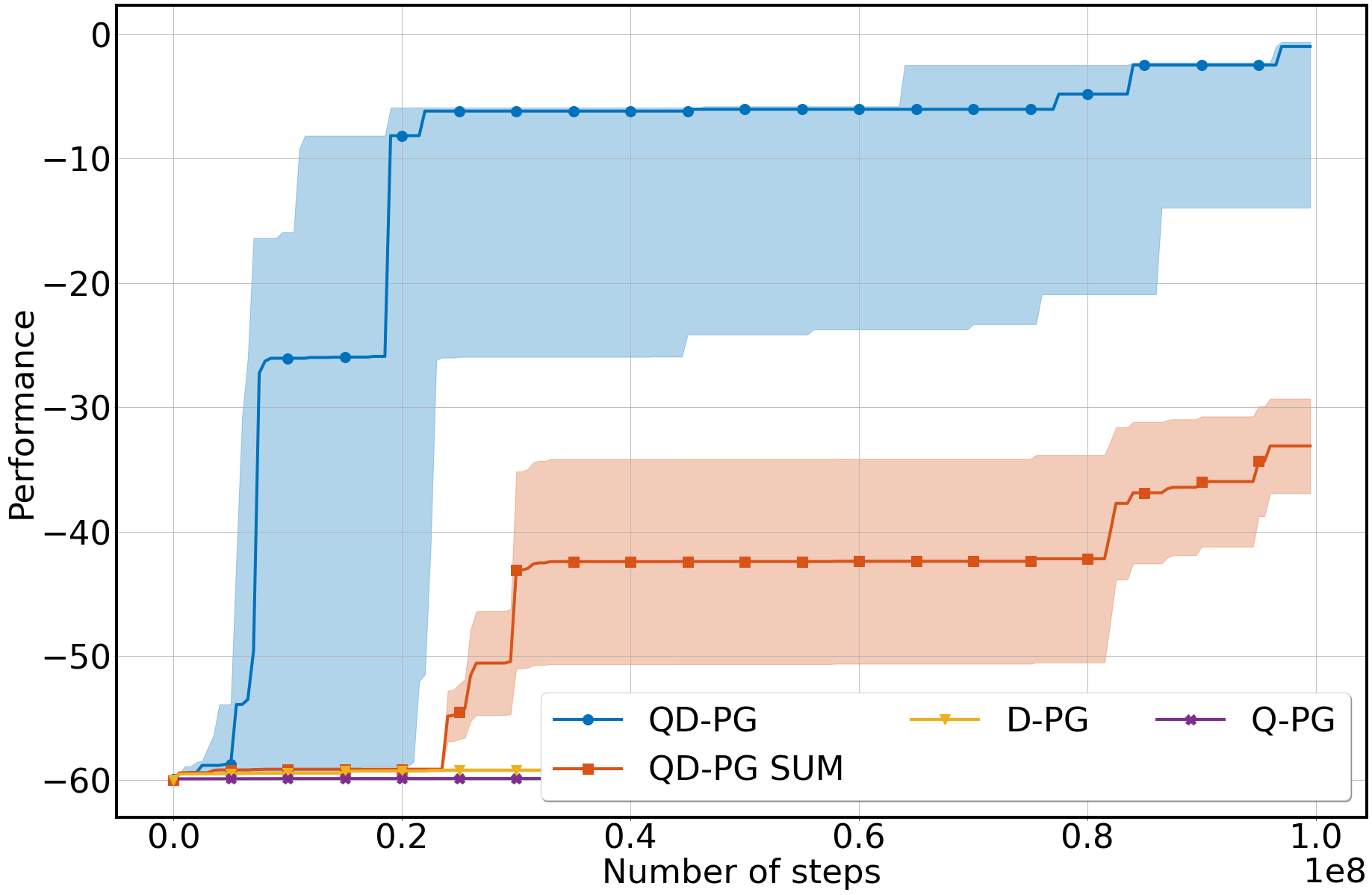}
  \caption{\antmaze: Ablation study}
  \label{fig:qdrlantmaze_ablation}
\end{subfigure}%
\begin{subfigure}{.5\linewidth}
\centering
  \includegraphics[width=1\linewidth]{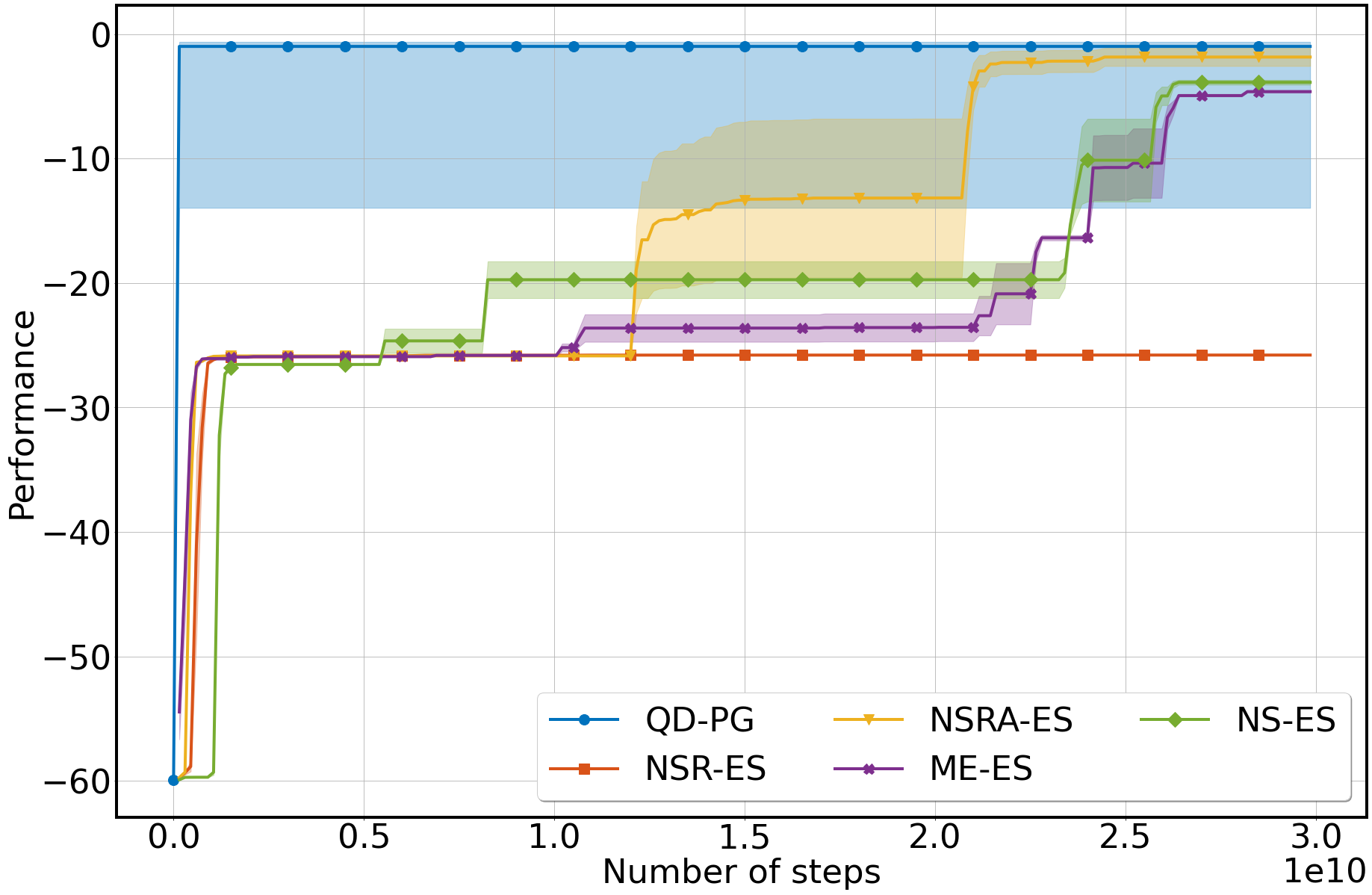}
  \caption{\antmaze: Evo methods}
  \label{fig:qdrlantmaze_evo}
\end{subfigure}
    \caption{Learning curves of \qdpg versus ablations and evolutionary baselines. In \ptmaze and \anttrap, the performance is the highest return. In \antmaze, it is the negative lowest distance to the goal. %In (c) and (d), curves show the best 2 seeds out of 5 for all algorithms.
    We separate the comparison on \antmaze into two graphs for better readability. Plots present median bounded by first and third quartiles.}
    \label{fig:results_charts_ablation_evo}
\end{figure}

Figure \ref{fig:results_charts_ablation_evo} shows supplementary results, with an ablation study and a comparison to evolutionary competitors in \antmaze. In \qdpg, the current population of solutions is evaluated every 150.000 time steps in \antmaze and \anttrap, and every 5000 time steps in \ptmaze. At evaluation time, agents are set to be deterministic and stop exploring. Figure~\ref{fig:results_charts_ablation_evo} reports the performance obtained by the best agent in the population at a given time step.

\begin{table}[h!]
    \centering
    \caption{Ablations and baselines summary. Selec. stands for selection. The last column assesses whether the method optimizes for a collection instead of a single solution.
    \label{tab:recap}}
    \begin{scriptsize}
    \begin{tabular}{lcccccc}
        \toprule
        %   & \textbf{Algorithm} & \textbf{Quality PG} & \textbf{Diversity PG} & \textbf{Quality Selection} & \textbf{Diversity Selection}\\
        & \textbf{Algorithm} & \textbf{QPG} & \textbf{DPG} & \textbf{Q Selec.} & \textbf{D Selec.} & \textbf{Collection}\\
        \midrule
        \multirow{4}{*}{\rotatebox[origin=c]{90}{Ablations}}
        & \qdpg & \checkmark & \checkmark & \checkmark & \checkmark & \checkmark\\
        & \qdpgsum & \checkmark & \checkmark & \checkmark & \checkmark & \checkmark\\
        & \dpg & X & \checkmark & \checkmark & \checkmark & \checkmark\\
        & \qpg & \checkmark & X & \checkmark & \checkmark & \checkmark\\
        \midrule
        \multirow{8}{*}{\rotatebox[origin=c]{90}{PG}}
        & \sac & \checkmark & X & X & X & X\\
        & \tddd & \checkmark & X & X & X & X\\
        & \rnd & \checkmark & \checkmark & X & X & X\\
        & \cemrl & \checkmark & X & \checkmark & X & \checkmark\\
        & \pssstddd & \checkmark & \checkmark & X & X & \checkmark\\
        & \agac & \checkmark & \checkmark & X & X & X\\
        & \diayn & X & \checkmark & X & X & X\\
        & \pgame & \checkmark & X & \checkmark & \checkmark & \checkmark\\
        \midrule
        \multirow{3}{*}{\rotatebox[origin=c]{90}{QD}} 
        & \mees & X & X & \checkmark & \checkmark & \checkmark\\
        & \nsres & X & X & \checkmark & \checkmark & \checkmark\\
        & \nsraes & X & X & \checkmark & \checkmark & \checkmark\\
        \bottomrule
    \end{tabular}
    \end{scriptsize}
\end{table}

\subsection{Coverage Maps}
\label{sec:supp_abl}
\begin{figure}[thbp!]
\centering
\begin{subfigure}{.47\linewidth}
\centering
  \includegraphics[width=1\linewidth]{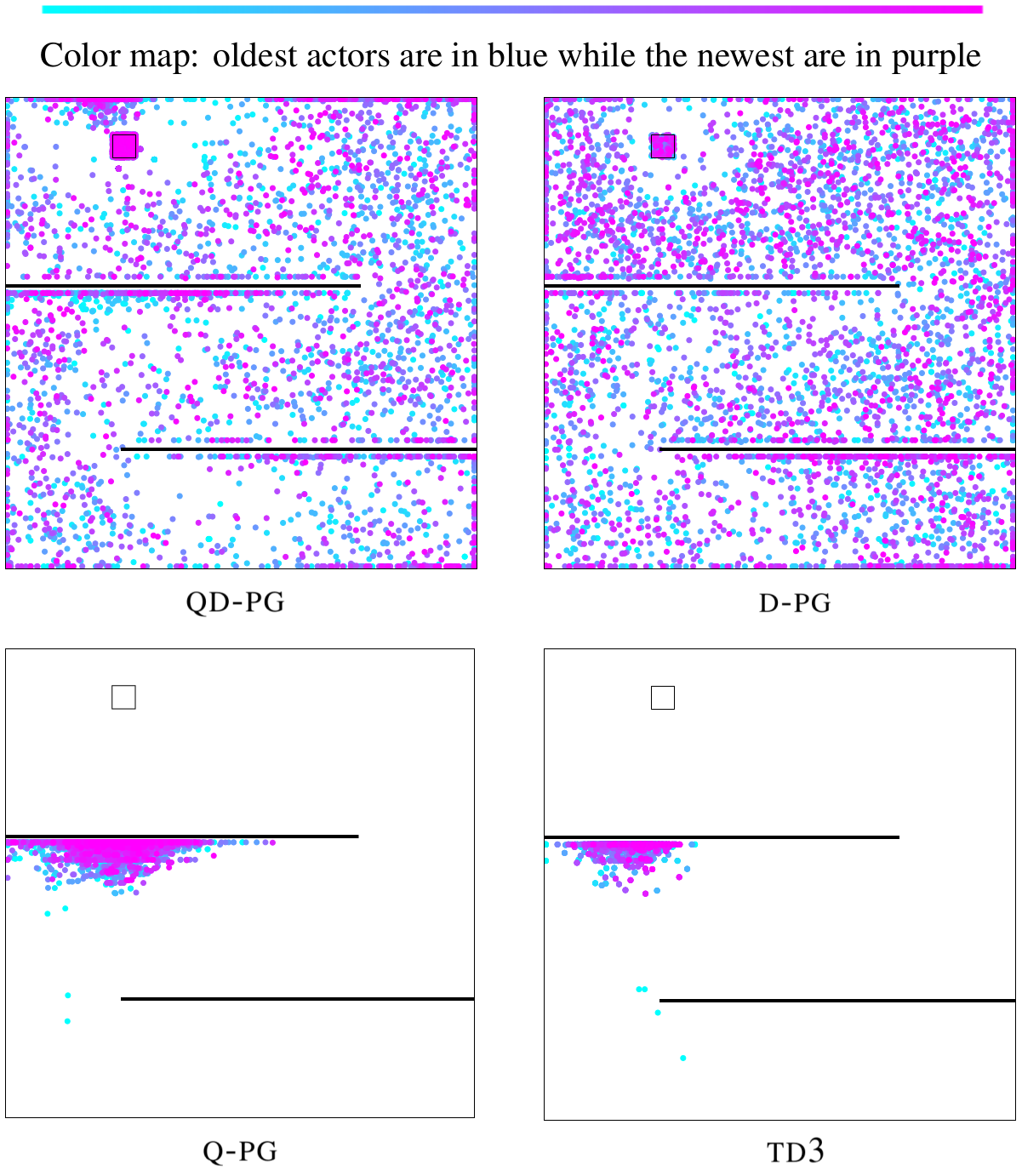}
  \caption{\ptmaze}
  \label{fig:coverage_map_pt_maze}
\end{subfigure}%
\hfill
\begin{subfigure}{.47\linewidth}
\centering
  \includegraphics[width=1\linewidth]{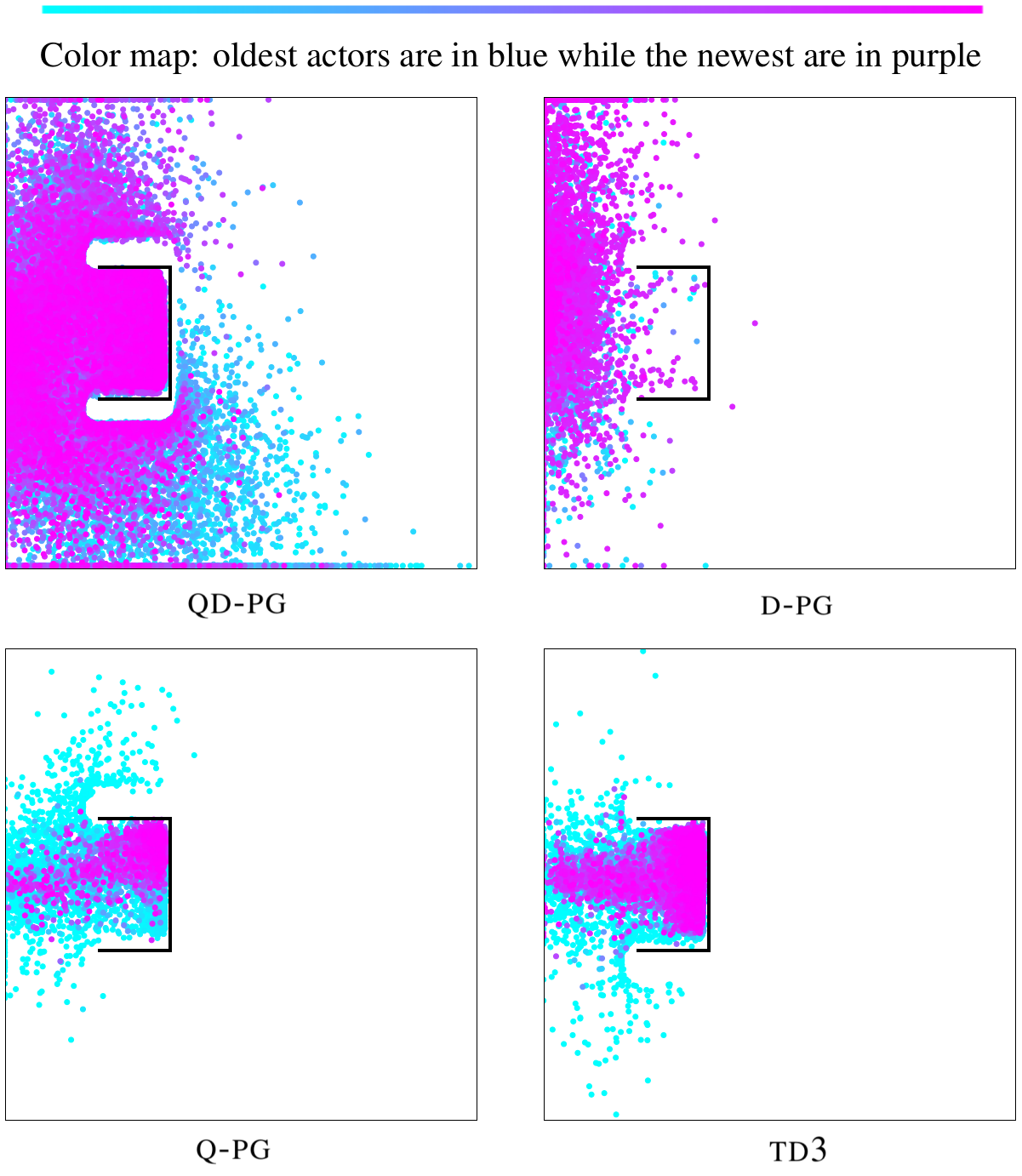}
  \caption{\anttrap}
  \label{fig:coverage_map_anttrap}
\end{subfigure}
    \caption{Coverage map of the \ptmaze and \anttrap environments for all ablations. Each dot corresponds to the final position of an agent.}
    \label{fig:coverage_maps}
\end{figure}

\figureautorefname{}~\ref{fig:coverage_map_pt_maze} shows coverage maps of the \ptmaze environment obtained with one representative seed by the different algorithms presented in the ablation study (see \tableautorefname{}~\ref{tab:comp_ablations_rl}). A dot in the figure corresponds to the final position of an agent after an episode. The color spectrum highlights the course of training: agents evaluated early in training are in blue while newer ones are represented in purple.

\qdpg and \dpg almost cover the whole BD space including the objective. Unsurprisingly, \qpg and \tddd present very poor coverage maps, both algorithms optimize only for quality and the \me selection mechanism in \qpg contributes nothing in this setting. By contrast, algorithms optimizing for diversity (\qdpg and \dpg) find the maze exit. However, as shown in \tableautorefname{}~\ref{tab:comp_ablations_rl}, \qdpg which also optimizes for quality, is able to refine trajectories through the maze and obtains significantly better performance.

\figureautorefname{}~\ref{fig:coverage_map_anttrap} depicts the coverage maps of the \anttrap environment by \qdpg and \tddd. Only \qdpg is able to bypass the trap and to cover a large part of the BD space.

\subsection{Fast adaptation}
\label{sec:supp_fast_adaptation}

The fast adaptation experiment described in Section \ref{sec:fast_adaptation} uses a Bayesian optimization process to quickly find a high-performing solution for a new randomly sampled goal.
Browsing the \me grid in an exhaustive way is another option to find a good solution for a new objective. However, the number of solutions to be tested with this option increases quadratically w.r.t. the number of meshes used to discretize the dimensions of the BD space.
As shown in \tableautorefname{}~\ref{tab:hyperparameters}, we use a $7 \times 7$ grid to train \qdpg in the \antmaze environment, containing a maximum of $49$ solutions. In this setting, the difference in computation cost between exhaustive search and Bayesian optimization is negligible.

\begin{figure}[thbp!]
    \centering
    \includegraphics[width=0.3\textwidth]{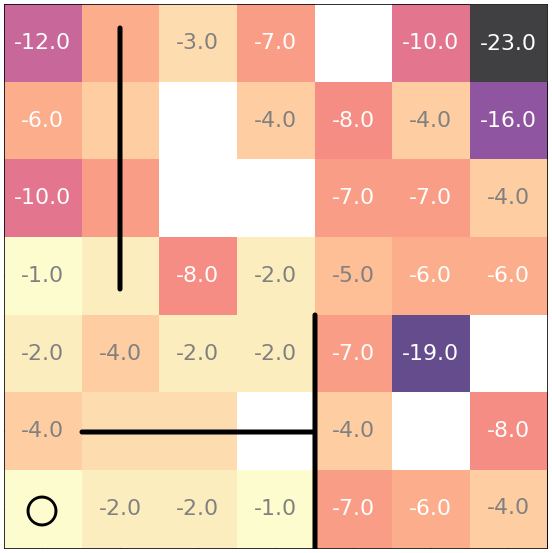}
    \caption{Performance map of 100 fast adaptation experiments in \antmaze. In each square, we display the score of the best experiment whose goal was sampled in this region of the maze, as several experiments may have goals in the same square. White squares correspond to regions where no goal was sampled during the experiments. The black circle shows the agent's starting position.}
    \label{fig:heatmap_fast_adaptation}
\end{figure}

To ensure that fast adaptation scales to finely discretized \me grids, we reproduce this experiment with a $100 \times 100$ grid, thus containing thousands of solutions. We first train \qdpg again on the standard objective of \antmaze and obtain a $100 \times 100$ grid of solutions. Then, we repeat the fast adaptation experiment described in Section~\ref{sec:fast_adaptation} using this large grid.
With a budget of only $50$ solutions to be tested during the Bayesian optimization process among the thousands of solutions contained in the grid, we are able to recover a good solution for the new objective.
We repeat this experiment $100$ times, each time with a new random goal, and obtain an average performance of $-9$ with a standard deviation of $7$.

Figure \ref{fig:heatmap_fast_adaptation} maps these $100$ fast adaptation experiments to their respective goal location and performance.
In each square, we display the score of the best experiment whose goal was sampled in this region of the maze.
For instance, the square in the top left corner of the performance map corresponds to one of the 100 fast adaptation experiments that sampled its goal in this part of the maze, and obtained a performance of $-12$ after testing $50$ solutions from the \me grid during the Bayesian optimization process.
Some squares do not have a score when no experiment sampled its goal in this region of the maze.

\section{\dpg derivations extended}
\label{sec:proof}

\subsection{Proof of Proposition 1.}
\begin{proof} (of proposition \ref{prop:div_gradient_1})

The {\em diversity} of a set of $K$ solutions $\{\theta_k\}_{k=1,\dots,K}$ is defined as $d: \Theta^K \rightarrow \mathbb{R}^+$:

\begin{equation}
    d\left(\{\mathbf{\theta}_k\}_{k=1,\dots,K}\right) = \sum\limits_{i=1}^K \min\limits_{k \ne i} ||\mathbf{\xi}(\mathbf{\theta}_i), \mathbf{\xi}(\mathbf{\theta}_k)||_{\mathcal{B}},
\end{equation}

The diversity can be split into three terms: the distance of $\theta_1$ to its nearest neighbor (defined as $\theta_2$), the distance of $\theta_1$ to the $\theta_j$ for which $\theta_1$ is the nearest neighbor (defined $\{\theta_j\}_{j=3, ..., K}$ \footnote{Remark: $\theta_2$ can appear twice in the list $\left(\theta_j\right)_{2 \leq j \leq J}$}) and a third term that does not depend on $\theta_1$. Thus:

\begin{align*}
     %$
     d(\{\theta_k\}_{k=1,\dots,K}) &= ||\xi(\theta_1), \xi(\theta_2)||_{\mathcal{B}} + \sum\nolimits_{j=3}^J ||\xi(\theta_1), \xi(\theta_j)||_{\mathcal{B}} + M \\
     &= \sum\nolimits_{j=2}^J ||\xi(\theta_1), \xi(\theta_j)||_{\mathcal{B}} + M
     %$,
\end{align*}
where $M = \sum\nolimits_{i \not\in \{1,\dots,J\}} \min\limits_{k \ne i} ||\xi(\theta_i), \xi(\theta_k)||_{\mathcal{B}}$ does not depend on $\theta_1$. Hence, the gradient with respect to $\theta_1$ is:

\begin{align*}
     %$
     \nabla_{\theta_1} d(\{\theta_k\}_{k=1,\dots,K}) = \nabla_{\theta_1} \sum\nolimits_{j=2}^J ||\xi(\theta_1), \xi(\theta_j)||_{\mathcal{B}}
     %$,
\end{align*}

As the remaining term is precisely defined as the novelty n :
\begin{align}
    n(\theta_1, \left(\theta_j\right)_{ 2 \leq j \leq J}) = \sum\nolimits_{j=2}^J ||\xi(\theta_1), \xi(\theta_j)||_{\mathcal{B}}
\end{align} 

We get the final relation :
\begin{align}
    \nabla_{\theta_1} d(\{\theta_k\}_{k=1,\dots,K}) = \nabla_{\theta_1} n(\theta_1, \left(\theta_j\right)_{ 2 \leq j \leq J})
\end{align} 

\end{proof}

\subsection{Motivation behind Equation~\ref{eq:diversity_gradient_constraint}}
The idea behind this equation is to link novelty defined at the solution level to a notion of novelty defined at the time step level. The information that we use at the time step level is the current state in the environment, so we describe a solution by the states it visits and hence want to define state novelty such that :

\begin{equation}
    n(\theta_1, \left(\theta_j\right)_{ 2 \leq j \leq J}) = \mathbb{E}_{\pi_{\theta_1}} \sum\limits_t n(s_t, \left(\theta_j\right)_{ 2 \leq j \leq J})
\end{equation}

Furthermore, we assume that a state is novel w.r.t. some solutions if it is novel w.r.t. to the states visited by these solutions. To be able to do so, we introduce the notion of state descriptor $\Psi$, which enables to define the novelty between states. Hence, we can define the novelty of a state w.r.t. to a set of solutions by: 

\begin{equation}
    n(s, \left(\theta_j\right)_{j=1,\dots,J}) = \sum ^J\limits_{j=1} \mathbb{E}_{\pi_{\theta_{j}}} \sum\limits_t ||\psi(s), \psi(s_t)||_{\mathcal{D}}
\end{equation}

This state descriptor $\Psi$ constrains the link between the novelty at the state level with the novelty at the solution level through a link with the behavior descriptor $\xi$.

That being said, we can now compute the diversity gradient thanks to the novelty at the state level. As a matter of fact, proposition \ref{prop:div_gradient_1} links diversity gradient to novelty at solution level:

\begin{equation}
    \nabla_{\theta_1} d(\{\theta_k\}_{k=1,\dots,K}) = \nabla_{\theta_1} n(\theta_1, \left(\theta_j\right)_{ 2 \leq j \leq J})
\end{equation}

Then we link novelty at the solution level to novelty at the state level with the following equation, which is satisfied thanks to the relevant choice of the state descriptor $\Psi$.
\begin{equation}
     n(\theta_1, \left(\theta_j\right)_{ 2 \leq j \leq J}) = \mathbb{E}_{\pi_{\theta_1}} \sum\limits_t n(s_t, \left(\theta_j\right)_{ 2 \leq j \leq J})
\end{equation}

Finally, by replacing the novelty at the solution level by the novelty at the state level in proposition \ref{prop:div_gradient_1}, we get the formulation of the diversity policy gradient given in equation~\ref{eq:dpg}.

\begin{equation}
    \nabla_{\theta_1} d(\{\theta_k\}_{k=1,\dots,K}) = \nabla_{\theta_1} \mathbb{E}_{\pi_{\theta_1}} \sum\limits_t n(s_t, \left(\theta_j\right)_{ 2 \leq j \leq J})
\end{equation}

\end{document}